\documentclass[manuscript,screen]{acmart}

\AtBeginDocument{%
  \providecommand\BibTeX{{%
    \normalfont B\kern-0.5em{\scshape i\kern-0.25em b}\kern-0.8em\TeX}}}

\setcopyright{acmcopyright}
\copyrightyear{2018}
\acmYear{2018}
\acmDOI{XXXXXXX.XXXXXXX}

\usepackage{tikz}
\usetikzlibrary{mindmap}

\DeclareMathOperator*{\argmax}{argmax}
\DeclareMathOperator*{\argmin}{argmin}
\DeclareMathOperator*{\cost}{cost}

\DeclareMathOperator*{\minimize}{minimize}
\DeclareMathOperator*{\softmax}{softmax}
\DeclareMathOperator*{\stateAttention}{\mathrm{\small{\texttt{state-attention}}}}
\DeclareMathOperator*{\actAttention}{\mathrm{\small{\texttt{action-attention}}}}
\newcommand{\ket}[1]{\lvert#1\rangle} % Ket
\newcommand{\norm}[1]{\left\lvert\left\lvert #1 \right\lvert\right\lvert} % vector/matrix norm

\usepackage[OT2,T1]{fontenc}
\usepackage{substitutefont}

\substitutefont{OT2}{\rmdefault}{wncyr}

\DeclareRobustCommand{\rn}[1]{% russian name
	{\fontencoding{OT2}\selectfont#1}%
}

\begin{document}

\title{Modeling Human Behavior Part II - Cognitive approaches and Uncertainty}

% \affiliation{%

\author{Andrew Fuchs}
\affiliation{%
  \institution{Universit\'{a} di Pisa, Department of Computer Science}
  \city{Pisa}
  \country{Italy}}
\email{andrew.fuchs@phd.unipi.it}

\author{Andrea Passarella}
\author{Marco Conti}
\affiliation{%
  \institution{Institute for Informatics and Telematics (IIT), National Research Council (CNR)}
  \city{Pisa}
  \country{Italy}
}

\renewcommand{\shortauthors}{Fuchs, et al.}

\begin{abstract}
As we discussed in Part I of this topic \cite{fuchs2022partI}, there is a clear desire to model and comprehend human behavior. Given the popular presupposition of human reasoning as the standard for learning and decision-making, there have been significant efforts and a growing trend in research to replicate these innate human abilities in artificial systems. In Part I, we discussed learning methods which generate a model of behavior from exploration of the system and feedback based on the exhibited behavior as well as topics relating to the use of or accounting for beliefs with respect to applicable skills or mental states of others. In this work, we will continue the discussion from the perspective of methods which focus on the assumed cognitive abilities, limitations, and biases demonstrated in human reasoning. We will arrange these topics as follows (i) methods such as cognitive architectures, cognitive heuristics, and related which demonstrate assumptions of limitations on cognitive resources and how that impacts decisions and (ii) methods which generate and utilize representations of bias or uncertainty to model human decision-making or the future outcomes of decisions.
\end{abstract}




\maketitle

% \tableofcontents

\section{Introduction}\label{sec:introduction}

% \todoInAP{Rewritten intro}
Future autonomous and adaptive systems are expected to further exploit the concept of cyber-physical convergence~\cite{CONTI20171}, and realize an environment where autonomous agents and humans work together as teams, understanding each other and anticipating each other's behavior and intentions. This is propelled, on the one hand, by the pervasive diffusion of connected devices in the physical environment, which are directly owned (e.g. personal devices) or in tight interaction (e.g. IoT devices) with the human user. On the other hand, the vast diffusion of AI can bring autonomy of agents to a new level, making their behavior much more refined, and adaptive to the varying conditions of the environment and users. Human-Centric AI (HCAI) is expected to be a fundamental element in this vision. Not only because users need to trust AI agents, e.g., thanks to explainable algorithm~\cite{10.1145/3236009}. Quite interestingly, AI agents will need to interpret human behavior in the context, so as to better interact with users, understand their actions, predict their choices, and ultimately orchestrate between actions performed directly by humans and those delegated to AI agents in autonomy. To this end, it is fundamental to equip AI agents with practical models of the human behavior. Potential application areas are numerous, spanning from robotics, medicine, e-health, autonomous driving, just to mention a few. A key distinction between the goals and approaches is often the fidelity of the replication and the expected deployment case. For instance, researchers may try to replicate the neurological pathways in an attempt to replicate the neuro-physical process underpinning reasoning \cite{asgaribrain}, or they may instead attempt to generate a computational model which is meant to mimic heuristically biased behavior \cite{lieder2017automatic}. In any case, a common aspect is the desire to use humans as the template for desirable patterns of reasoning.

Though emerging, the literature on HCAI and related human behavioral models is quite vast already. Various types of HCAI exist. A first kind of HCAI systems focus primarily on explainability, making sure users can understand the process leading to a certain outcome by the AI agent~\cite{yang2020re, eiband2021support}. A more advanced form of HCAI (typically referred to as hybrid intelligence) consists in AI agents and humans interacting directly, and impacting each other's operations~\cite{gurcan2021mapping, kaluarachchi2021review}. Examples of such cases are AI agents learning in presence of human teachers~\cite{puig2020watch, liu2019self, ramaraj2021unpacking, navidi2021new, najar2021reinforcement, holzinger2019interactive}, or humans exploiting the outcome of AI agents to acquire better comprehension of a phenomenon~\cite{schneider2020humans}. Moreover, in other cases humans and AI agents perform a common operation as a team~\cite{reddy2018shared, morrison2021social}. In this context, designing approaches to orchestrate delegation of tasks and decisions among the team's members is also fundamental~\cite{ning2021survey, raghu2019algorithmic}.

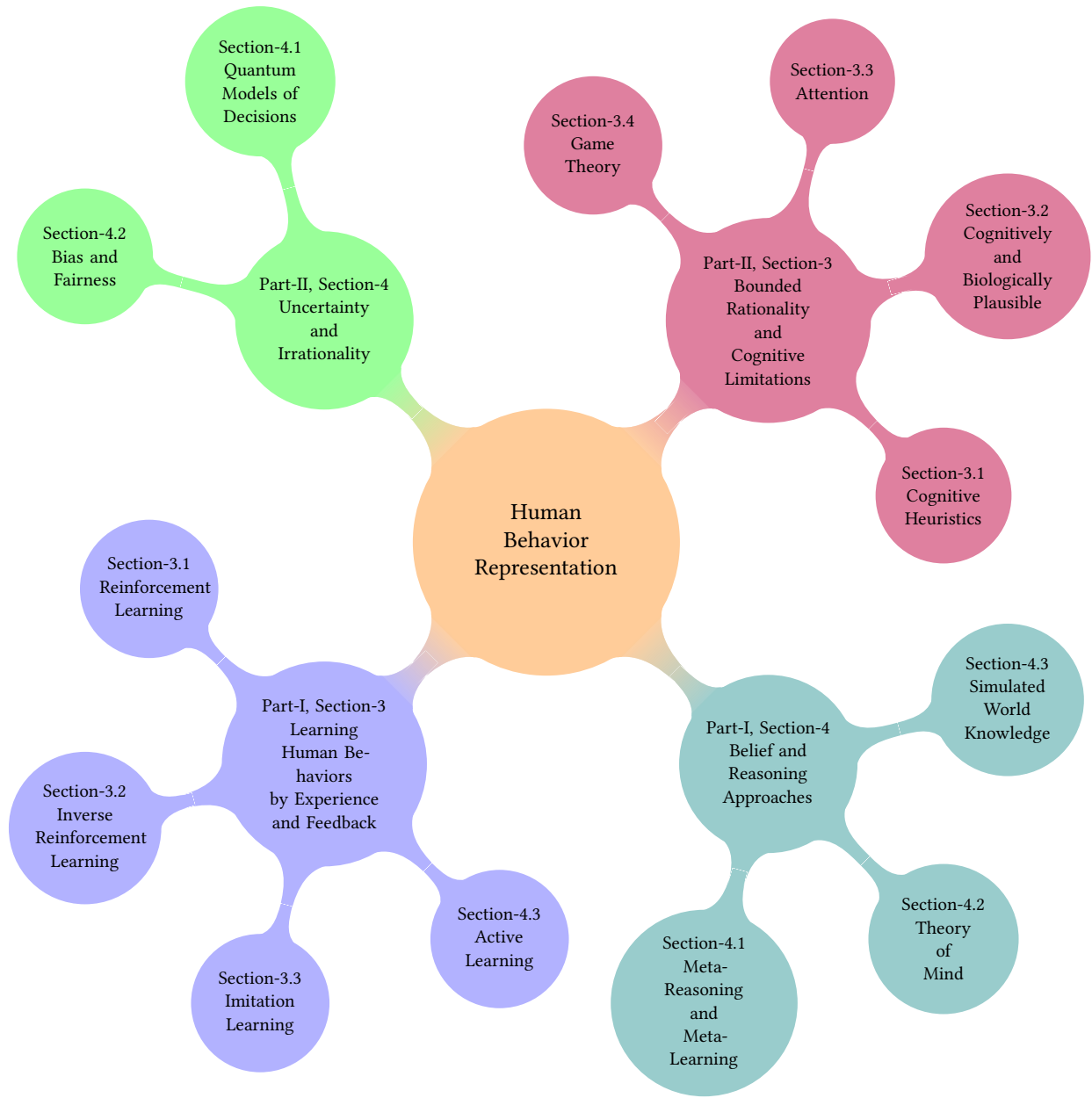
\begin{figure}
    \begin{tikzpicture}[mindmap, grow cyclic, text width=2.5cm, inner sep=1mm, every node/.style=concept, concept color=orange!40,
        level 1/.append style={level distance=4.75cm,sibling angle=90},
        level 2/.append style={level distance=3.75cm,sibling angle=60}]
    
    \node{Human\\Behavior\\Representation}
        child [concept color=blue!30, font=\fontsize{8pt}{10pt}\selectfont] { node {Part-I, Section-3\\Learning\\Human Behaviors\\by Experience\\and Feedback}%Modeling\\and\\Learning\\Behavior}
            child [font=\fontsize{8pt}{10pt}\selectfont]{ node {Section-3.1\\Reinforcement\\Learning}}
            child [font=\fontsize{8pt}{10pt}\selectfont]{ node {Section-3.2\\Inverse\\Reinforcement\\Learning}}
            child [font=\fontsize{8pt}{10pt}\selectfont]{ node {Section-3.3\\Imitation\\Learning}}
            child [font=\fontsize{8pt}{10pt}\selectfont]{ node {Section-4.3\\Active\\Learning}}
        }
        child [concept color=teal!40, font=\fontsize{8pt}{10pt}\selectfont] { node {Part-I, Section-4\\Belief and\\Reasoning\\Approaches}
            child [font=\fontsize{8pt}{10pt}\selectfont]{ node {Section-4.1\\Meta-Reasoning\\and\\Meta-Learning}}
            child [font=\fontsize{8pt}{10pt}\selectfont]{ node {Section-4.2\\Theory\\of\\Mind}}
            child [font=\fontsize{8pt}{10pt}\selectfont]{ node {Section-4.3\\Simulated\\World\\Knowledge}}
        }
        child [concept color=purple!50, font=\fontsize{8pt}{10pt}\selectfont] { node {Part-II, Section-3\\Bounded Rationality\\and\\Cognitive Limitations}
            child [font=\fontsize{8pt}{10pt}\selectfont]{ node {Section-3.1\\Cognitive\\Heuristics}}
            child [font=\fontsize{8pt}{10pt}\selectfont]{ node {Section-3.2\\Cognitively\\and\\Biologically\\Plausible}}
            child [font=\fontsize{8pt}{10pt}\selectfont]{ node {Section-3.3\\Attention}}
            child [font=\fontsize{8pt}{10pt}\selectfont]{ node {Section-3.4\\Game\\Theory}}
        }
        child [concept color=green!40, font=\fontsize{8pt}{10pt}\selectfont] { node {Part-II, Section-4\\Uncertainty\\and\\Irrationality}
            child [font=\fontsize{8pt}{10pt}\selectfont]{ node {Section-4.1\\Quantum\\Models of\\Decisions}}
            child [font=\fontsize{8pt}{10pt}\selectfont]{ node {Section-4.2\\Bias and\\Fairness}}
        };
    \end{tikzpicture}
    \caption{Taxonomy of concepts}
    \label{fig:taxonomy_of_concepts}
\end{figure}

Figure~\ref{fig:taxonomy_of_concepts} provides a taxonomy of the approaches existing in the literature to address these aspects. Work falling in the areas of direct learning of human behavior and modeling beliefs and reasoning are covered extensively in~\cite{fuchs2022partI}. Specifically, they deal with approaches to learn human behavior by trial-and-error (such as Reinforcement Learning or Instance-based Learning) and approaches to prescriptively describe key features of human behavior such as reasoning processes and beliefs into a model (such as Theory of Mind). In this paper (Part II) we complete the survey by presenting in detail approaches falling in the other two classes depicted in Figure~\ref{fig:taxonomy_of_concepts}.

In this paper, we will discuss some of the popular topics and applications relating to Human-Centric AI, Human-AI Interaction, and Hybrid Intelligence. These topics include: bounded rationality and heuristics, cognitively plausible representations, bias, and more. These topics represent methods which attempt a model mimicking or inspired by differing biological/neurological, cognitive, and social levels of reasoning. Additionally, we will discuss how these topics align with application areas of interest. These application areas cover a wide assortment of both scenario as well as level of autonomy expected. For instance, this can include topics such as demographic preferences \cite{jackson2017agent} to something as safety-critical as fully autonomous driving \cite{wu2020joint, fernando2020deep}. The specific scenario can rely significantly on the level of autonomy expected and the level of risk or control humans are willing to allow. In Section~\ref{sec:bounded_rationality_and_cog_limits}, we discuss methods which attempt to replicate the cognitive abilities and sub-optimality in humans. These allow for models which replicate biased or bounded rational behavior inspired by human cognition. Key to this Part Is modeling human cognitive resources, and ways adopted by humans to efficiently use them, possibly at the cost of obtaining imprecise understanding of the learned process or make mistakes. In Section~\ref{sec:uncertainty_and_irrationality} we focus on approaches that model uncertainty in the human reasoning process, to the point of leading to choices that do not appear to be the outcome of a rational process. Finally, Section~\ref{sec:conclusions} provides a critical discussion on the approaches presented in the paper. Before presenting them, we briefly mention key application areas in Section~\ref{sec:applications_and_related}, highlighting how the considered approaches can be applied there. Note that, as in~\cite{fuchs2022partI}, we present each specific approach according to a common scheme. First, we point to specific surveys and related dealing in greater detail with that topic. Then, we discuss the general principles. Next, we discuss one concrete example where those principles are made practical. Finally, we briefly mention additional examples where the same principles have been applied.

Finally, also in this paper, as in~\cite{fuchs2022partI}, we limit the analysis to (a representative subset of) works providing \emph{quantitative} models (e.g. math models or algorithms), as these are the approaches that allow to ``code" human behavior in autonomous systems.

\section{Samples of Application Areas and Related}\label{sec:applications_and_related}

In this section, we will briefly discuss popular application areas demonstrating uses of the techniques discussed in this paper. This list is not comprehensive, but serves to demonstrate topics which are likely more familiar and of immediate interest. The approaches used demonstrate methods which serve to replicate, model, or learn from human behavior and capabilities.

\subsection{Driver Prediction and Autonomous Driving}

There have been numerous examples of research performed to model and predict behavior in a driving scenario \cite{kolekar2021behavior}. In the case of autonomous vehicles, there is a need to model and predict the outcome of control delegation between the autonomous system and the human driver. To do so, researchers have investigated cognitive models to predict the time to take over control given the type and difficulty of actions the human driver is performing when the control is switched \cite{lotz2020take, scharfe2019cognitive, scharfe2019towards}. This allows for a model which can simulate the cognitive and bio-mechanical responses when changing tasks for the human drivers.

\subsection{AI in Games and Teaching}

In the area of video and serious games, AI is being considered with respect to multiple aspects. In Part I of this topic, we discussed approaches integrating AI methods for learning models of behavior. In the case Serious Games, systems can be implemented to train human users and learn models of their behavior \cite{khan2020serious}. This allows for systems which can teach and allow for improvement with a more thorough understanding of the user's behavior. For instance, cognitive architectures (see Section~\ref{sec:cog_architectures}) can be used to define intelligent tutoring systems \cite{streicher2021dynamic}.

\subsection{Agent-Based Modeling}

To achieve a model of human behavior and decision-making, numerous topics have been investigated. Some examples include agent-based modeling \cite{jackson2017agent, kennedy2012modelling, groeneveld2017theoretical, singh2018behavior, dobson2019integrating}, which allows for models of groups of people or populations. In the case of agent-based modeling, models of human behavior are often defined and then studied in an environment over a simulated timeline. The agents follow the defined patterns of behavior and the resulting global patterns can be analyzed. For instance, population segregation based on demographic preferences regarding neighbors can be modeled by defining a diversity preference and modeling the movement of agents in an environment \cite{jackson2017agent}. Populations of humans can also be modeled with other techniques designed to model the interdependence of the agents. Interactions can be modeled mathematically with humans represented as nodes in a network, particles in an environment, or more \cite{dolfin2017modeling}. Additionally, the model of human behavior in an ABM can also be supported by models of Uncertainty, Bounded Rationality, or Cognitive Architectures, which we discuss in Section~\ref{sec:bounded_rationality_and_cog_limits} and Section~\ref{sec:uncertainty_and_irrationality} with topics such as Section~\ref{sec:cog_architectures}, Section~\ref{sec:game_theory}, etc. Moreover, As we will focus on more direct and individual-level models of behavior, we will not present this topic in greater detail in this paper.

\section{Bounded Rationality and Cognitive Limitations}\label{sec:bounded_rationality_and_cog_limits}

In the following sections, we will discuss methods inspired by the cognitive limitations and characteristics demonstrated in human reasoning. The approaches discussed in Section~\ref{sec:cog_heuristics} will demonstrate how humans utilize heuristics to enable fast and frugal reasoning as well as more deliberative systems. This combination of systems enables more efficient use of cognitive resources and has been demonstrated in numerous studies of human reasoning. Related to these concepts, we see the use of systems designed to replicate the cognitive and neurological performance (not necessarily the physical structure) seen in humans (see Section~\ref{sec:cog_architectures}). These systems are designed to mimic human performance on tasks by replicating how humans use knowledge and memories to make decisions and perform tasks. Next, in Section~\ref{sec:attention}, we discuss techniques which are inspired by how humans attend to stimuli and make associations between observed values. This can relate to human vision (i.e. foveation) or how humans identify correlations between different items in the same context (e.g. the word `book' in a sentence would increase the relevance of the word 'library' in the same sentence). Next, in Section~\ref{sec:quantum_representation}, we will discuss the use of quantum representations to accommodate for uncertainty and provide methods which can support a quantum representation of states. Last, we will outline concepts relating to bounded rationality in the context of Game Theory in Section~\ref{sec:game_theory}. Similar to previous sections, Section~\ref{sec:bias_and_fairness}) will discuss topics which focus less on a direct replication of cognitive functions; instead, these topics focus on the resulting biases that come from the use of heuristics and similar shortcuts in reasoning. Further, these limitations in reasoning generate immediate and long-term effects motivating studies on fairness.

\subsection{Cognitive Heuristics}\label{sec:cog_heuristics}

\subsubsection{Relevant survey(s)}

For relevant survey papers and related, please refer to \cite{booch2020thinking}

\subsubsection{Principles and Definitions}

Bounded rationality describes the notion of humans making rational choices under the constraints ascribed to cognitive limitations of the decision-maker \cite{simon1990bounded, rizun2014simulation, askari2019behavioral}. These constraints are a reflection of the assumed limitations or deficiencies in a human's computational abilities/capacity and knowledge. Similarly, these constraints can be viewed as respecting a notion of cognitive or computational cost. As a reflection of these limitations, there is an assumption that humans perform decision-making in a manner which allows them to find a reasonable approximation of the optimal solution while reducing overall cost or time. Reasonable could be viewed as a ``good enough" or \emph{satisficing} solution. The process of finding such an acceptable result is supported by shortcut techniques (i.e. heuristics) which allow a person to approximate the collection of alternative solutions by finding a set of satisfactory alternatives.

An additional aspect of bounded rationality is the assumption that humans will sacrifice exactness or optimality of a solution for the sake of efficiency. In this context, efficiency can refer to time required to a solution, cognitive resources needed, etc. In general, this alludes to a sense of frugality when it comes to the cognitive resources a human is willing to dedicate to a decision process. In most circumstances, this frugality does not cause issues and allows for sufficient stimulus processing. On the other hand, there are examples of where this may cause a significant omission of perception. For example, humans can be tasked with observing a scene and then asked questions at the end of the observation pertaining to specific content. Often, there can be items or people hidden in plain sight due to the observer being distracted by more attention-grabbing stimuli, which demonstrates possible errors in the use of cognitive heuristics by the human brain \cite{wang2013invisible}. 

\paragraph{Dual-System Reasoning}\label{subsubsec:dual_system}

Relating to bounded rationality, there has been extensive interest in what is referred to as dual-system reasoning or dual process theory, now extending to topics in AI \cite{booch2020thinking}. The argument is that humans utilize two systems of reasoning based on the context of the problem and the limitations of their cognitive systems. The assumption is the two levels handle the problems at different speeds, fidelity, cognitive cost, etc. These distinctions are based on the belief that humans tend to utilize a lower-cost reasoning system when the penalty for a sub-optimal solution is minor or when the time or cognitive burden of reaching a higher accuracy solution is too great (see \cite{peterson1967man, milli2021rational} for more on cognitive cost). For simplicity, we follow a common convention and refer to them as System-1 and System-2. For an example of System-1 reasoning, catching a falling object typically doesn't leave sufficient time for deeper reasoning, so we rely on instinctive movements made quickly by System-1. On the other hand, System-2 can support deeper reasoning, longer time to a decision, etc.

System-1 is commonly assumed to be based on approximations generated via heuristics. These heuristics provide shortcuts to reasonably accurate solutions. For example, it has been argued that humans utilize what is known as the availability heuristic, which selects a solution based on the strongest association between the current situation and memories of potentially similar instances. This means humans will tend to place higher weight on memories more closely aligned with the current observations. As such, they will be biased toward solutions with higher likelihood of recall. This allows humans to use similar past experience to simplify the decision-making process when using System-1. If the problem is too complex for System-1, then System-2 needs to be utilized. System-2 allows for deeper inspection and the possibility to use or combine multiple underlying processes.

The use of these systems can lead to biases in reasoning and potential incorrect assessments. A simple example is the Gambler's Fallacy, which demonstrates how humans tend to believe that a sequence of flips from a fair coin should be self-correcting \cite{kahneman1971belief, tversky1974judgment}. In other words, when the coin is flipped multiple times, a sequence of identical outcomes is considered less and less likely as the length of the sequence grows. This misconception leads the person to feeling that the alternate outcome should be more likely in the next instance. As is apparent from the independence of the samples, this is in fact incorrect reasoning. Probability theory dictates that the outcomes of each toss should have no effect on the next toss. However, this is a common bias observed in human behavior resulting from this form of reasoning.

Humans demonstrate additional forms of heuristic-based and biased reasoning. The following are some additional examples \cite{facione2012think}:
\begin{itemize}
	\item Satisficing: Use a sufficient option rather than the optimal one
	\item Affect: Make decision based on intuition or ``gut feeling"
	\item Simulation: Estimate likelihood of an outcome based on the how easy it is to imagine the outcome
	\item Availability: Estimate the likelihood of a future event based on the strength of recall for similar past occurrences
	\item Representation: Assume X is the same as Y when you notice X is similar to Y in some way(s)
	\item Association: Connect ideas based on the word association and the memories, meanings, or impressions they trigger
	\item Optimistic Bias: The tendency to underestimate our own risks and overestimate our own control in dangerous situations
	\item Hindsight Bias: The tendency to remember successful events as resulting from your own decisions and failures as resulting from bad luck or decision from others
	\item Loss and Risk Aversion: Avoid risk and loss by maintaining the status quo
	\item All or Nothing: Simplify decisions by treating remote probabilities as if they were not even possibilities
\end{itemize}
These heuristics/biases demonstrate systems which can generate correct solutions in many cases, but can also lead to misconceptions or ignored information. As a result, it stands to reason that methods which wish to model the behavior and reasoning of humans will need to take these potential inaccuracies into account.

\subsubsection{Applications and Recent Results}

In \cite{lieder2017automatic}, the authors demonstrate an application of the Dual-Process concept of human reasoning and use of heuristics in a RL paradigm demonstrating multi-alternative risky choice in the Mouselab scenario (widely used to study decision strategies). In this scenario, their approach demonstrates the emergence of two known heuristics: take-the-best (TTB) (chooses alternative favored by the most predictive attribute and ignores others) and random choice. They note these are resource-rational strategies for low-stakes decisions with high and low dispersion of their outcome probabilities, respectively. They further note how the TTB heuristic is commonly used by humans when under time pressure and one outcome is much more likely than others. Similarly, the authors demonstrate how humans tend to accept random selection when the stakes are low in low-dispersion cases. The authors represent the bounded optimal decision process as a meta-level Markov Decision Process by considering the cost of computing a solution which impacts the utility of a decision or action. The actions are treated as costly computations, necessitating the ability to make decisions with efficiency in mind. This need for efficiency follows those seen in the justification of the representation and use of heuristics.

An augmentation to the MDP considered in this research is the meta-level MDP. In this case, actions for the meta-level MDP are cognitive operations $\mathcal{C}$ performed in belief states $b_t\in\mathcal{B}$. Additionally, the meta-level MDP has a transition function $T_{meta}$ and reward function $r_{meta}\in R_{meta}$. The operations in $\mathcal{C}$ include an operator $\perp$ which terminates deliberation and subsequently translates the current belief into an action. The determination to end deliberation and select an action can be seen as a representation of how humans select System-1 or System-2 reasoning, which then results in an outcome from the selected system. The reward $r_{meta}$ combines the cognitive cost $c\in\mathcal{C}$ with the expected immediate reward the agent expects to receive once deliberation terminates and an action is taken. In the case of a computation, the reward is defined as $r_{meta}(b_t,c)= -\cost(c)$ for $c\in\mathcal{C}$; otherwise, $r_{meta}(b_t, \perp) = \argmax_a b_t^{(\mu)}(a)$ where $b_t^{(\mu)}(a)$ is the expected reward of action $a$ according to belief $b_t$.

\paragraph{Results}

The MouseLab scenario provides a testbed in which agents can improve the likelihood of successful decisions by performing additional information acquisitions. While the acquisition improves the decision, it also incurs a cost. Therefore, the agent should minimize the occurance of cognitive costs while maximizing the subsequent game outcome. This promotes a tradeoff of decision quality and decision time, mimicking the similar processes witnessed in human cognition. The authors note their proposed method rediscovered Take-The-Best (TTB), Weight-Additive Strategy (WADD), and then random choice strategy. The additional strategy, WADD, is performed by computing the expected values of all gambles using all possible payoffs.

There were three noted outcomes regarding the predictions and the pattern which justify use of heuristics and match the observation of study participants ($200$ participants on Amazon Mechanical Turk). First, the model predicted fast-and-frugal heuristics should be prioritized/utilized more frequently in high-dispersion trials (high dispersion means an outcome significantly more likely than the others and fast-and-frugal heuristics ignore all outcomes except the most probable). Second, the model indicates the utility of simple heuristics, primarily when the stakes are low. Third, the model indicates the benefit of increased time and effort for high-stake scenarios to receive the highest possible payoff.

\paragraph{Additional Relevant Results}

There have been numerous studies and examples demonstrating the errors in human reasoning and many of the studies relate the failings to this form of reasoning. Some examples include: \cite{lau2001advantages, lee2007representation, tversky1974judgment}. Additionally, researchers have used heuristics to represent how information diffusion occurs in a network of individuals \cite{mordacchini2020human, mordacchini2017social, conti2013design, mordacchini2016design, arnaboldi2017online}.

\subsection{Cognitively/Biologically Plausible Representations}\label{sec:cog_architectures}

\subsubsection{Relevant survey(s)}

For relevant survey papers and related, please refer to \cite{kotseruba202040}

\subsubsection{Principles and Definitions}

With the goal of achieving a general AI (i.e. reaching human-level intelligence \cite{lieto2018role}), there have been numerous approaches inspired by the cognitive mechanisms enabling the intelligence observed in humans. In \cite{russell2002artificial}, the authors note several ways how reaching human-level general intelligence might be possible. One noted method relates to the design and justification of cognitive architectures. In the case of cognitive architectures, the goal isn't always to achieve a perfect analog of the human brain and its neurological function; instead, a common goal is to generate a system capable of demonstrating the same kinds of abilities and deficiencies seen in human cognition, reasoning, intuition, etc. (e.g. perception, memory, attention)\cite{kotseruba202040, thomson2015general, dimov2020model, gonzalez2003instance, kelly2019high}. Under these circumstances, the goal is often the creation of a model of behavior which fits the cognitive/neurological dynamics of the human brain \cite{kotseruba202040, anderson2013adaptive, whitehill2013understanding, ritter2019act, asgaribrain, sun2012reasoning, urban2001pecs, sinz2019engineering}.

As noted in Section~\ref{subsubsec:dual_system}, it is generally accepted that humans reason with systems operating at different levels of fidelity. Humans can make faster and cognitively frugal decisions or utilize slower and more cognitively burdensome resources. As such, research has been dedicated to the creation of systems demonstrating these characteristics (and beyond) \cite{kotseruba202040}. These systems demonstrate an ability to learn behavior as we've seen in previous sections in Part I (e.g. Reinforcement Learning), but the distinction in this case is the emphasis on replicating the cognitive performance of humans. This distinction motivated us to place a higher emphasis on the cognitive and biologically plausible mechanisms of this portion of the paper. Further, there have been studies which show these representations can provide the best performing (and likely best fitting) approximations to human cognitive performance \cite{stocco2021analysis}. For example, cognitive architectures utilize memory systems which can replicate how humans retain information and utilize that information when making decisions. As a result, we see this section as more suitable in a cognitive limitations and biases context.

\paragraph{World Representation - Symbolic, Emergent, and Hybrid}

To support reasoning and behavior, the system needs a method for representing the world. For cognitive architectures, there are three main categories for the underlying representations: symbolic, emergent, and hybrid. As the name would suggest, symbolic systems use symbols to represent concepts or knowledge. Given the symbols, the system can manipulate them by using a given set of instructions. The instructions can be provided through if-then rules or similar means. As can be expected, a symbolic representation allows for accurate planning and reasoning., but the potential downside being that this approach is brittle and does not adapt to changes in the environment. Emergent systems operate similar to what is seen in Artificial Neural Network (ANN) systems. Information is processed by the system and associations are made through a learning process. This of course increases the system's responsiveness to changes in an environment, but can reduce the transparency or the easiness of interpreting the system's behavior. To utilize the advantages of both systems, with the hope of overcoming the shortcomings, there are hybrid systems which combine the symbolic and emergent approaches.

\paragraph{Learning Methods}

Learning in a cognitive architecture can be performed in several ways including: Declarative, Procedural, Associative, etc. \cite{kotseruba202040}. In the case of Declarative learning, the system is provided a collection of facts about the world as well as relationships between them. For instance, many systems such as ACT-R, SAL, CHREST, or CLARION utilize chunking mechanisms to declare new knowledge items. For Procedural learning, the system learns skills gradually through repetition, which can be accomplished through the accumulation of examples of successful execution of a task or problem. More closely aligned to RL, Associative learning is based on observations of rewards or punishments.

\paragraph{Memory}

Architectures can be supported by different memory mechanisms depending on the type of capabilities being replicated. When performing a task, the memory utilized to temporarily store information related to the task at hand is referred to as \emph{working memory} \cite{anderson1996working}. This memory is updated rapidly as the state of the world changes and actions are taken. Further, there is commonly an assumption regarding the capacity limitations of working memory for humans. In addition to working memory, other systems provided a means to accomplish long-term memory storage. This can support storage of procedural memory to define basic skills or behavior or declarative memory for knowledge. This allows for the storage of innate skills as well as accumulated knowledge. Additionally, some systems are defined with a hybridization of long and short-term memory, referred to as \emph{global memory}. This results in all knowledge and memories being represented by the same system.

\subsubsection{Applications and Recent Results}

The above characteristics are broad aspects covering different approaches for cognitive architectures. For a specific example, we present a recent result based on the ACT-R architecture. The authors present a cyber security game designed to demonstrate cognitive biases of cyber attackers \cite{cranford2020toward}. This displays how humans are susceptible to fallacies in reasoning which result in sub-optimal and biased behavior. The authors demonstrate how their models replicate the biases motivating human behavior patterns in system selection and the choice to abandon a system and forfeit the previous effort on the current system. 

The authors used an Instance-based Learning (IBL) model using the Adaptive Character of Thought - Rational (ACT-R) architecture. ACT-R is a theory of cognition which models how humans recall “chunks” of information from memory and how humans solve problems by splitting them into sub-goals \cite{whitehill2013understanding}. Knowledge is applied from working memory as needed to find a pattern of behavior meeting the goal. This model utilizes techniques designed to mimic human memory retrieval, pattern matching, and decision making. IBL uses ACT-R's blending mechanism, which interpolates across past experiences to estimate an outcome. The interpolation is weighted by the contextual similarity between the present observation or instance and the past experiences. This provides an estimate expected outcome based on the \emph{consensus value} $V$ which minimizes the dissimilarity (measured by $Sim$) from the values contained in instance $i$ defined as:
\begin{equation}\label{eqn:cog_arch_application_value}
    \argmin_V\sum_i P_i\times (1 - Sim(V,V_i))^2
\end{equation}
where $i$ refers to an instance stored as a memory chunk representing a past state-action-outcome observation and $P_i$ refers to the retrieval probability (based on IBL-based measures). In the case where $Sim$ is interpreted as the error, then Equation~\ref{eqn:cog_arch_application_value} generates a least-squared error method \cite{lebiere1999dynamics}. In other words, this finds an estimated value $V$ which best fits the past observations that are weighted by their strength of recall. These estimated values are used to make a determination regarding which action/production should be executed. The measure or threshold which determines whether an action is available for execution limits the set of possible actions further. This means that the value is based on a representation which considers how strongly a memory is remembered, how similar the memory is to the current context, and the value observed by the choice made in that past observation. The strength of a memory represented by the retrieval probability utilizes a Boltzmann softmax equation
\begin{equation}
    P_i = \frac{e^{A_i/\tau}}{\sum_j e^{A_j/\tau}}
\end{equation}
where $\tau$ defines the temperature parameter, which scales probabilities defined by the activation function. The activation function provides a measure of how strongly a memory is remembered and associated with the current context. This strength is based on elapsed time since the observation was made. The activation for a chunk or instance $i$ is defined as
\begin{equation}\label{eqn:cog_arch_application_activation}
    A_i = \ln{\sum_{j=1}^n t^{-d}_j + MP\times\sum_k Sim(\nu_k,c_k) + \epsilon_i}
\end{equation}
where $t_j$ refers to the elapsed time since the $j^{th}$ occurrence of instance $i$, $d$ is the decay rate (commonly set to $0.5$), $c_k$ refer to the context elements, $\nu_k$ refer to the instance in memory, and $MP$ is the mismatch penalty (in this case, set to the default of $1.0$). The first term in Equation~\ref{eqn:cog_arch_application_activation} provides the measure of strength based on the time elapsed and the second term is another similarity term similar to what is seen in Equation~\ref{eqn:cog_arch_application_value}. $MP$ is a weight term parameter which scales the similarity scores in the sum, and the last term, $\epsilon_i$, is a variance parameter providing stochasticity in the activation function. Similar to Equation~\ref{eqn:cog_arch_application_value}, the $Sim$ measure ensures the memories considered are a suitable match to the current context in order to prevent consideration of too dissimilar of instances. In more general terms, the above equations define a method for determining which memories are considered, how strongly they impact the estimate based on past observations, and how the resulting behavior occurs based on this historically weighted knowledge.

The agents are trained to perform the cyber attacker role. As such, the agents are provided observation instances which include the probability of a system being monitored, the reward for successful infiltration of a system, the penalty for detection, and a warning signal denoting whether a system is being monitored. The model is then primed with seven instances: five simulating a practice round, two representing knowledge of occurrences (absent and success, absent and uncertainty). This provides the system with an initial set of experience to allow for initialization of learning behavior without relying on random decisions. The model then is trained for four rounds of $25$ trials.

\paragraph{Results}

The model was tested in comparison to human performance. Human players were studied to generate a baseline of behavior and identify any demonstrated biases in outcomes. Based on the experiments, the authors show the human players demonstrating preferences or likelihoods of attack for different systems. They also demonstrated the cognitive systems performing equivalent preferences/probabilities.

\paragraph{Additional Relevant Results}

As mentioned, an underlying justification for using cognitively plausible approaches is the use of algorithms which mimic the performance of humans under the same task. Through the use of cognitive architectures, the cognitive and bio-mechanical processes for a human attempting to accomplish a goal can be replicated. For instance, models of user swipe behavior, mental folding, driver takeover of an autonomous vehicle, simulated cyber attackers, pilots, etc. can be created \cite{russwinkel2018act, scharfe2019towards, cranford2020toward, cranford2020adaptive, cranford2020cognitive,  preuss2019implementation, klaproth2020neuroadaptive, lotz2020take, klaproth2019act, scharfe2019cognitive}. The structure of the artificial systems architecture can have a significant impact on performance \cite{schrimpf2020neural} and there are numerous examples of architectures and their underlying structures \cite{kotseruba202040}.

Recent research has investigated closer ties between cognitively inspired representations and topics in RL. For example, there have been attempts to utilize these representations to observe and predict behavior of agents in an environment \cite{nguyen2020cognitive, nguyen2020effects}. These demonstrate the use of an agent's ability to learn a model of likely behavior in an attempt to anticipate likely next steps. The observer sees actions selected by RL agents and builds a model of likely behavior, be it next actions or likely goal states. Another aspect of mimicking processes exhibited by humans is the ability to imagine. This is demonstrated in \cite{zhu2008daydreaming} where they create an agent which can perform conceptual blending and daydreaming in order to generate dialogues inspired by the user behavior. Similarly, \cite{driessens2003relational} demonstrates a similar use of related past instances and object relations in an RL context.

For a simulation environment supporting these sorts of representations, \cite{smart2016integrating} demonstrates an integration of the ACT-R architecture and Unity 2D/3D modeling system. This allows for control and learning for artificial agents in a Unity environment with the behavior being managed by ACT-R representations. This enables testing of cognitively inspired representations of behavior in a simulated world, which can enable the use of advanced game engine features for fidelity of the world model. In a similar topic, \cite{pentecost2016using} demonstrates integrating a physics engine with a cognitive architecture to aid an agent's decision making. This is shown to aid the agent in playing racquetball in a simulated environment.

Cognitive representations can also be utilized to define an agent intended to interact or coordinate with a human. For example, \cite{morita2020cognitive} utilizes a cognitive architecture and RL concepts to predict and model behaviors in a HAI context. The scenario involves control of a moving circle on a screen which needs to follow a path which is scrolling down the screen. The human can either choose to take control or pass off control to the agent.

In another context, \cite{lansdell2019learning} demonstrates training of an RL agent using a sort of layered approach which they refer to as a two-learner system. The layering in their approach refers to the use of two agents to learn a policy. The first agent is used to learn an approximation of feedback signal which would result from backpropagation and the second learns the behavior policy based on the approximated feedback. The argument is the two-learner system could simulate cortical neuron physiology.

\subsection{Attention}\label{sec:attention}

\subsubsection{Relevant survey(s)}

For relevant survey papers and related, please refer to \cite{correia2021attention, graziano2019attributing, niv2019learning}.

\subsubsection{Principles and Definitions}

Humans and other living beings do not process all the available perceptual information available to them. Instead, they utilize a cognitive and behavioral system which allows them to reduce the complexity of perception through an objective or subjective selectiveness with respect to information. This selectivity or bias is referred to as attention \cite{correia2021attention}. A basic interpretation justifying the biological need for attention would be the fact that our environment provides more stimuli than we can reasonably process fast enough for our environment. In this case, fast enough is with respect to the actions or behaviors necessary for survival.

When facing a critical situation, timeliness can be crucial; otherwise, an overload of stimuli could cause a costly delay (e.g. moving out of the path of an oncoming vehicle). As such, our brains allow us to reduce, or even ignore, information perceived in order to reduce the cognitive burden. Further, attention allows us to prioritize the information and assign more or less significance based on learned or perceived importance. In the oncoming vehicle example, it is likely not important to note the color of a building in the distance while estimating the speed and trajectory of the vehicle.

From a computational perspective, attention initially was studied primarily from the context of vision \cite{correia2021attention} where images were studied with under the task of identifying salient regions of the image. Artificial systems were developed to generate maps which would filter the input for processing. With the growing popularity of Deep Learning (DL), attention techniques were transitioned to neural network paradigms. Attention is used to modify the flow or processing of information in the network(s). The task of learning attention allows the systems to learn how to ignore stimuli, similar to the natural analogs mentioned. This allows systems to contextually alter the significance of information in order to better suit the underlying task. The attention mechanisms utilized in DL can be categorized as follows:

\paragraph{Attention}

Soft attention uses softmax functions to weight the input elements with a weight value in $(0,1)$. This allows the system to learn and utilize an interdependence between different input parameters. Being based on softmax functions, soft attention provides a differentiable mechanism for attention. The soft attention scales the relative intensity of the input parameters.

\paragraph{Hard Attention}

Hard attention, as the name suggests, is the complement of soft attention. It utilizes weights in $\{0,1\}$ to generate a mask to signify whether information is used or entirely ignored. As a result, the hard attention mechanism is non-differentiable. This necessitates a learning process for determining where to assign the weight values. In this case, there is a distinct exclusion of regions of the input domain while the remainder is observed at normal scale.

\paragraph{Self-Attention}

In self-attention, the system is learning an interdependence between sequential input elements. This allows a system to identify and utilize a notion of relation between items in the same input sequence. As a result, self-attention can be useful in understanding deeper relationships between items in the input rather than a holistic view of the input. For example, \cite{vaswani2017attention} introduces the transformer network which performs self-attention using representations of queries $Q$, keys $K$ and values $V$
\begin{equation}
    \textrm{Attention}(Q,K,V) = \softmax\left(\frac{QK^T}{\sqrt{d_k}}\right)V
\end{equation}
where $d_k$ refers to the dimension. In this case, the attention mechanism learns associations between different components of the input and their corresponding strengths via a representation as keys $K$ and queries $Q$. This allows the model to learn a relationship between the current task, the input data, and the current query. The associations are learned as weight matrices which scale the input values and give the weighted strength of association. This generates a weighted association between elements of the input and forms a compatibility measure of the values.

\subsubsection{Applications and Recent Results}

In \cite{oroojlooy2020attendlight}, the authors demonstrate the use of attention with a RL agent to control simulated traffic lanes. The motivation for the use of attention provided references the fact that traditional systems would require retraining for a new lane configuration. The use of attention allows for more flexible representations and can handle different numbers of roads/lanes. The proposed algorithm is tested against several baselines and demonstrates strong performance in traffic regulation.

For a road with intersection $m\in\mathcal{M}$, define the \emph{traffic characteristics} $s^t_l$ of lane $l\in\mathcal{L}$ at time $t$, where $\mathcal{L}$ denotes the set of all approaching lanes to the intersection. Additionally, $\mathcal{L}^{in}$ and $\mathcal{L}^{out}$ refer to entering and exiting lanes respectively, and so $\mathcal{L} = \mathcal{L}^{in}\bigcup\mathcal{L}^{out}$. Further, define \emph{traffic movement} $\nu_l$ as a set that maps traffic of lane $l\in\mathcal{L}^{in}$ to possible leaving lanes $l'\in\mathcal{L}^{out}$. In this context, the set of valid traffic movement $p\in\mathcal{P}$ during a green light are called a \emph{phase}. Define participating lanes $\mathcal{L}_p$ as the set of lanes that have appeared in at least one traffic movement of phase $p$. Note that each phase has a minimum time and following that time, a decision about the next phase should be made.

The definition of the RL problem requires translating the domain into states, actions, and rewards. They utilize the traffic characteristic $s^t_l$ as the state at time $t$. For actions, this is represented by the assigned active phase at time $t+1$. Authors use two attention mechanisms to define the policy $\pi^t$ to select the next action at time $t$, or better the next phase. Specifically, the first attention mechanism defines weights to be used in considering the states of the corresponding lanes. This attention step, modelled in Equation~\ref{eqn:attention_first}, generates a vector of weights $w^t_p\coloneqq\{w^t_l, l\in\mathcal{L}_p\}$, where each weight corresponds to a specific lane relevant for phase $p$ (i.e., $l\in\mathcal{L}_p$). The weights depend on a function ($r^t_p$) of the current states ($s^t$) of these lanes, and the average of the function across all relevant lanes, $q^t_p$. Intuitively, this attention allows to focus the mechanism on lanes depending on their closeness to the ``average" (relevant) lane.
\begin{equation} \label{eqn:attention_first}
    w^t_p = \stateAttention(r^t_p,q^t_p),\forall p\in\mathcal{P}
\end{equation}

The weights are used to compute a representation of each phase $p$, as $z^t_p = \sum_{l\in\mathcal{L}_p} w^t_l\times g(s^t_l)$. The representations of the possible phases $p\in\mathcal{P}$ are then fed into an LSTM, to capture the sequential dependence between phases. The output of the LSTM ($o^t$) is used, together with the representation of the possible phases $z^t_p,\, p\in\mathcal{P}$ for the second attention mechanism, modelled as in Equation~\ref{eqn:attention_second}. Specifically, this provides the probability $\pi^t\coloneqq\{\pi^t_p,p\in\mathcal{P}\}$ of switching to any of the phases of the next time step.
\begin{equation} \label{eqn:attention_second}
    \pi^t = \actAttention(\{z^t_p,p\in\mathcal{P}\},o^t)
\end{equation}
The reward is based on the negative of the intersection pressure defined in \cite{wei2019presslight}. The intersection pressure relates the lane capacity and the lane flow:
\begin{equation}
    w(l,m) = \frac{\nu_l}{\nu^{max}_l} - \frac{\nu_m}{\nu^{max}_m}
\end{equation}
for incoming and outgoing lanes $l$ and $m$, where $\nu^{max}_l$ refers to the maximum lane capacity for lane $l$. Note that this is an indication of the incoming and outgoing flow of traffic. Based on this paradigm, the RL agent learns a policy $\pi$ which suggests a phase for the next time-step for the current state $s^t$. This policy is learned based on the algorithm illustrated above and uses the cumulative rewards for policy updates. This provides the means to map states to actions in the AttendLight algorithm.

\paragraph{Results}

The AttendLight algorithm was tested using three-way and four-way intersections with varying numbers of lanes, phases, and flow rates. The states $s^t_l$ represent chunks $c\in\{1,2,3\}$ of the road leading up to the intersection for lane $l$ and are $100$ meter segments of a $300$ meter length of the lane. Each lane $l$ has a corresponding number of vehicles $\alpha^t_{l,c}$ in a chunk $c$ at time $t$. Further, lanes also may contain waiting vehicles and the quantity of waiting is represented by $\beta^t_l$. Therefore, the traffic characteristic for a lane is defined as $s^t_l\coloneqq[\alpha^t_{l,1},\alpha^t_{l,2},\alpha^t_{l,3},\beta^t_l]$. The proposed algorithm is tested against a number of reference baselines in the literature, showing a significant improvement in terms of lower Average Travel Time (ATT)

\paragraph{Additional Relevant Results}

Additional examples of interesting use cases for attention include \cite{ferret2019self}, which attempts to address self-attention in RL to help solve credit assignment. Similarly, \cite{ke2018sparse} utilizes attention to address the temporal credit assignment problem. Further, \cite{vaswani2017attention, zhao2019explicit} are in reference to the widely popular transformer network. The work in \cite{vaswani2017attention} provides an introduction of the approach while \cite{zhao2019explicit} extends the approach to introduce sparsity to allow the most contributive components for attention to be reserved while the other irrelevant information are removed. This is effective in preserving important information and removing noise, so the attention can be much more concentrated. Further, \cite{kerg2020untangling} demonstrate the use of in an LSTM/RNN context for copying, transfer, and denoising tasks.

\subsection{Game Theory}\label{sec:game_theory}

\subsubsection{Relevant survey(s)}

For relevant survey papers and related, please refer to \cite{samuelson1995bounded, muller2017towards}. Note that, since game theory is a widely investigated and well-known topic, in the following we only sketch very briefly the key ideas behind this theory. The main reason for mentioning it in this survey is to place it in the overall context of human behavioral models. Also, we don't mention explicitly applications and results, as the literature is huge and pinpointing only a (few) specific example(s) would be not that useful for the readers.

\subsubsection{Principles and Definitions}

Extending and utilizing the notion of rational behavior, game theory focuses on the interdependence of choices when the circumstances involve a collection of individuals \cite{muller2017towards}. In game theory, the decision-maker, often referred to as the player, is operating with an assumed feedback signal. The feedback (payoff) provides a value associated with how desirable or costly an outcome might be for the player. Based on the payoff and anticipated behavior of other players, a player can attempt to optimize their choice (action) in order to best ensure an acceptable outcome. In this case, rather than attempting to learn a policy of behavior, the models of utility define the constraints of an optimization problem. Based on these constraints, the players of a game can find a suitable solution which follows the assumptions of rationality.

More formally, a player $i$ will have a set of available actions $A$ from which they can select an action $a_i$. The optimality of an action is dependent on multiple factors. First, the player has a payoff function $u : A \rightarrow \mathbb{R}$ that will map an action to a value. This mapping depends on the definition of the problem and the interdependence of actions to values between the players. The values $u(a)$ given by $u$ allow the player to generate a preferential ordering of actions, which indicates a player's need to identify desirable actions. Therefore, if the preferences of the player satisfy the following, then that player is considered rational under certainty \cite{askari2019behavioral}:

\begin{enumerate}
	\item (Completeness) $a_1 \succeq a_2$ or $a_2 \succeq a_1$
	\item (Transitivity) If $a_1 \succeq a_2$ and $a_2 \succeq a_3$, then $a_1 \succeq a_3$
\end{enumerate}

The payoff values and the assumptions relating to rationality can then be utilized to identify an appropriate action or behavioral policy. In game theory, the behavioral policy of a player is referred to as the player's \emph{strategy} or \emph{strategy profile}.

The strategy of a player is a distribution over actions indicating the likelihood of selecting an action. A strategy which places the mass on more than one action would be considered a mixed strategy, which indicates a player does not select a particular action $100\%$ of the time in a given scenario (i.e. a pure strategy).

The identification of a strategy is the process of finding an equilibrium. In an Nash equilibrium, the player strategies are such that deviation from the current strategy for any player would be undesirable as it would lead to another player having the means to take advantage of the change in order to achieve a better result. Depending on the game, it is possible to find zero, one, or even multiple equilibria. In the case of multiple equilibria, it is possible for the payoff for a particular player to vary between the equilibria, but it remains true that a deviation from the equilibrium from a single player would be undesirable \cite{muller2017towards}.

Based on the problem definition, the approach for finding an equilibrium often comes down to an optimization problem. Given a set of $N$ players $\mathcal{N}=\{1,\dots,N\}$ where each player $\nu$ has their own strategy $x^\nu\in\mathcal{X}_\nu\subseteq\mathbb{R}^{n_\nu}$, each player has an objective function $\theta_\nu (x^\nu,x^{-\nu})$ and constraints $g^\nu_i (x^\nu,x^{-\nu}) \leq 0,  (i=1,\dots,m_\nu)$, which depend on their strategy and the strategy of others $x^{-\nu}\coloneqq (x^{\nu'})_{\nu'\neq \nu}$, an equilibrium in a Nash Game can be found as the solution to the optimization problem \cite{kim2021equilibrium}:
\begin{align}
     \minimize_{x^\nu} \quad & \theta_\nu(x^\nu,x^{-\nu})\nonumber\\
    P_\nu (x^{-\nu}) \quad \textrm{subject to} \quad & g^\nu_i (x^\nu,x^{-\nu}) \leq 0, i=1,\dots,m_\nu\nonumber\\
    & x^\nu \in \mathcal{X}_\nu\nonumber
\end{align}
The payoffs and assumption of rationality provide the constraints for an optimization-based solution method. However, the interdependency of the behaviors, outcomes, and payoffs can increase the difficulty of finding a suitable solution.

\section{Uncertainty and Irrationality}\label{sec:uncertainty_and_irrationality}

\subsection{Quantum Representations of Decisions and Irrational Thinking}\label{sec:quantum_representation}

\subsubsection{Relevant survey(s)}

For relevant survey papers and related, please refer to \cite{jones2020cerebral, moreira2018quantum, dehdashti2020irrationality, aerts2021modeling, marcot2019advances}

\subsubsection{Principles and Definitions}

It has been argued that traditional probabilistic representations do not fully represent the reasoning of humans \cite{moreira2019towards} or can require exponentially more complex representations \cite{moreira2018quantum, marcot2019advances}. Instead, researchers have suggested the use of quantum-based methods for representing the statistical/probabilistic relationships between knowledge \cite{jones2020cerebral}. The argument is that the superposition-like representation better demonstrates how humans can have varying beliefs, which might not directly match the assumptions or requirements of probability (e.g. summing to one). This allows for a representation which can perform in cases where the reasoning currently operates in a state where multiple outcomes are possible or represent the same indefinite state. This is also a proposed method to account for potentially irrational or probability-violating reasoning of humans \cite{dehdashti2020irrationality, huang2019uncertainty}. The use of quantum representations also allows for replicating or modeling fallacies in human reasoning \cite{aerts2021modeling}. A common method for this representation is the use of a quantum-based Bayesian Network. In this case, a similar representation of the Bayesian network is utilized, but the dynamics are represented using quantum representations of probabilities.

\paragraph{Quantum Dynamics for Decision Models}

As defined in the Reinforcement Learning section of Part I, sequential decision making can be modeled using a MDP. This representation provides a mechanism for representing the transition between world states based on a decision or action. Such a representation enables learning associations between actions and outcomes to build a model of behavior. In \cite{busemeyer2006quantum}, the authors suggest the use of a quantum dynamics model to represent brain processes as a replacement for the classic MDP model. The quantum representation allows for transitioning from a `single path' assumption to one which represents unknown previous states as a `superposition of states'. The superposition model allows for modeling interference effects for unobserved paths, which violates the Markov representation.

As noted by \cite{busemeyer2006quantum}, a key distinction in the representations is as follows:
\begin{quote}
    According to the Markov model, for any given realization, the unobserved system occupies exactly one basis state $\ket{j}$ at each moment in time. A sample path of the Markov process is a series of jumps from one basis state to another, which moves like a bouncing particle across time. A different path is randomly sampled for each realization of the Markov process. According to the quantum model, for any given realization, the unobserved system does not occupy any particular basis state at each moment in time. A realization of the quantum process is a fuzzy spread of membership across the basis states, which moves like a traveling wave across time. Each realization of the quantum process is identical, producing the same series of states across time. All of the randomness in the quantum model results from taking a measurement.
\end{quote}
Hence, the quantum representation supports the notion of wave interference in the dynamics representation.

From a modeling standpoint, the main difference between a Markov and a Quantum representation is therefore the following. In a \emph{specific realization} of a Markov process, the state at any point in time is deterministic and can be represented with $\ket{P(t)}$ as follows
\begin{equation}
    \ket{P(t)} = \sum_{j\in\Omega} w_j(t)\cdot\ket{j},
\end{equation}
where $j$ denotes the possible states, $\Omega$ is the set of possible states, and $w_j(t)\in\{0,1\}$ is an indicator variable. Essentially, $\ket{P(t)}$ provides the (unique) state of the process at time $t$ for the specific realization.

On the contrary, in a quantum representation, the specific realization, at any point in time $t$ is not deterministically in one and only one state, but is in a \emph{superposition} of all possible states, each with a given \emph{weight}. In other words, even for a specific realization the process can be in \emph{any} of the possible states, while the uncertainty can be only removed by ``measuring" the specific state of the process. Specifically, the state, at time $t$ of the realization, denoted as $\ket{\psi(t)}$ is modeled as follows
\begin{equation}
    \ket{\psi(t)} = \sum \psi_j(t)\cdot\ket{j}    
\end{equation}
where where $\psi_j(t)$ is a complex number representing the probability amplitude that the process is in the specific state $\ket{j}$ at time $t$. As in any quantum superposition case, note that the squared magnitude of $\psi$ must be unity (i.e. $|\psi|^2=1$) in order to ensure the squared amplitudes produces a probability distribution over the possible states.

Finally, transitions between states become, in the case of quantum representation, specific quantum operations over the representation of the states provided by $\ket{\psi(t)}$, which translate, in the quantum domain, the traditional concept of transition probabilities between states of deterministic Markov Processes.

\subsubsection{Applications and Recent Results}

In \cite{he2018evidential}, the authors investigate the use of quantum representations in the context of categorization. The use of a quantum model is intended as a means to represent interference effects observed in categorization and the resulting impact of decision making. The inclusion of a quantum system allows the modeling of a state which represents uncertainty in the reasoning process. This models how humans demonstrate hesitance when facing a decision. For their experiments, the authors  investigate two paradigm conditions: categorization decision-making (C-D) and decision alone (D alone). In the C-D condition, in each trial, participants were shown pictures of faces, which vary along two dimensions: face width and lip thickness. Participants were asked to categorize the face as a “good” ($G$) guy or a ``bad" ($B$) guy and then make a decision to ``attack" ($A$) or to ``withdraw" ($W$). In the D alone condition, the participants were asked to make a decision directly without categorizing, but the faces shown were the same as in the C-D condition.

\paragraph{Proposed Method}

Although categorization happens in the belief representation, it can influence the action part by producing the interference effect, which can also model the disjuction fallacy (i.e. the false judgement that the probability $P(A \cup B)$ is less than either $P(A)$ or $P(B)$). Consequently, the authors utilized a method to predict it. For the proposed problem, the initial state involves six combination of beliefs and actions
\begin{equation}
    \{\ket{B_GA_A},\ket{B_GA_U},\ket{B_GA_W},\ket{B_BA_A},\ket{B_BA_U},\ket{B_BA_W}\}
\end{equation}
where, for example, $\ket{B_GA_A}$ symbolizes a participant categorizing the face as good while still intending to attack. Since participants are assumed to have some potential to be in any of the six quantum states, the person's state is a superposition of the six basis states:
\begin{align}
    \ket{\psi} = &\psi_{AG}\cdot\ket{B_GA_A} + \psi_{UG}\cdot\ket{B_GA_U} + \psi_{WG}\cdot\ket{B_GA_W}\\\nonumber
    &+ \psi_{AB}\cdot\ket{B_BA_A} + \psi_{UB}\cdot\ket{B_BA_U} + \psi_{WB}\cdot\ket{B_BA_W}
\end{align}
with initial state corresponding to an amplitude distribution:
\begin{equation}
    \psi(0) = 
    \begin{bmatrix}
        \psi_{AG}\\
        \psi_{UG}\\
        \psi_{WG}\\
        \psi_{AB}\\
        \psi_{UB}\\
        \psi_{WB}
    \end{bmatrix}
\end{equation}
where $|\psi_{XY}|^2$ is the probability of observing state $\ket{B_XA_Y}$ initially.  As an assumption, the the initial state is treated as equally distributed. In the process of decision making, updated information regarding a player’s beliefs causes a transition in belief states. The decision-maker must convert this transition in reasoning states into a decision.
To convert the observations and measurements into actions, participants must convert the uncertain state $U$ to either $A$ or $W$. This represents when the decision maker transitions from uncertainty due to hesitation to a decision state. In this case, the authors propose the use of Pignistic Probability Transformation (PPT). This provides the following total probability of attacking given that the face is categorized as $G$ or $B$ respectively:
\begin{equation}
    P(A|G) = \norm{\Psi(A|G) + \frac{1}{2}\Psi(U|G)}^2
\end{equation}
\begin{equation}
    P(A|B) = \norm{\Psi(A|B) + \frac{1}{2}\Psi(U|B)}^2
\end{equation}
where $\Psi(A|X)$ is the conditional amplitude (i.e. quantum equivalent of conditional operator) of attacking given the face is categorized as $X\in\{G,B\}$. The above denotes an even attribution of the probability of transitioning from the uncertain state to decision. This then provides the means to calculate the total probability of attacking:
\begin{align}
    P(A) &= P(G)P(A|G) + P(B)P(A|B)\nonumber\\
         &= \sum_{X\in\{G,B\}}\left(|\psi_{AX}|^2 + |\psi_{UX}|^2 + |\psi_{WX}|^2\right)\cdot\norm{\Psi(A|X) + \frac{1}{2}\Psi(U|X)}^2
\end{align}

A similar approach is utilized for $P(W)$ and in the D alone condition. This allows a model of representing the state of uncertainty encountered before a decision is made and the process of converting updated beliefs into an action. Based on this model, the action likelihoods can be represented and estimated using a quantum modeling approach.

\paragraph{Results}
For experiment conditions, the proposed method was tested against prior observations of human-generated data as well as prior prediction model results. In the data generation process, $26$ participants were asked to categorize the face as \emph{good guy} ($G$) or \emph{bad guy} ($B$) and then make a decision to \emph{attack} ($A$) or to \emph{withdraw} ($W$). The faces roughly fall into two categories: “narrow” faces (narrow width and thick lips) or ``wide" faces (wide width and thin lips). Participates were informed ``narrow" faces had a $0.6$ probability of belonging to the ``bad guy" population. Similarly, participants were informed “wide” faces had a $0.60$ probability of belonging to the “good guy” population. Rewards were were given for choosing attack for $B$ and withdraw for $G$. In the D alone condition, the participants were asked to make a decision directly without categorizing, but the faces shown to participants were the same as those in the C-D condition. Each participant provided $51$ observations for the C-D condition and  $17$ observations for the D alone condition.  The proposed approach showed a strong alignment with the pattern of behavior observed in the human participant results.

The authors also demonstrate the performance of the proposed approach with respect to sensitivity analysis. The authors show how sensitive the result of the method are to a $\pm 5\%$ change in the original human results. In this test, the method again shows strong performance, but the authors note a stronger sensitivity to changes in the $B$ case versus the $G$ case. For a plot of these results, please refer to the original article.

\paragraph{Additional Relevant Results}

In addition to the prior examples, quantum representations can also support fuzzier associations between knowledge items. The use of quantum representations allows for deviations from normative probabilistic methods and can support a notion of interference between variables and paradoxical behavior \cite{moreira2020quantum, huang2019uncertainty}. When considering multiple individuals, quantum-based methods have also been investigated to model the diffusion of information \cite{mutlu2020quantum}. The flexibility of representation from quantum methods also enables handling of the disjunction fallacy [i.e. the false judgment that $P(A \cup B)$ is less than either $P(A)$ or $P(B)$], which violates the law of total probability \cite{he2018evidential}. Quantum representations can also be used to represent states in game theory settings where there is uncertainty about the potential behavior of others or the current state \cite{pothos2009quantum, moreira2018quantum}. Further, as noted above, the concept of Bayesian Networks and causality can be extended to utilize representations supported by quantum dynamics \cite{moreira2020quantum, dehdashti2020irrationality, huang2019uncertainty}.

\subsection{Biases and Fairness in Representations and Understanding}\label{sec:bias_and_fairness}

\subsubsection{Relevant survey(s)}

For relevant survey papers and related, please refer to \cite{gigerenzer2008moral, oaksford2001probabilistic, dasgupta2020algorithmic, suomala2020consumer, booch2020thinking, dasgupta2020algorithmic, mandelbaum2020can, kennedy2012modelling, lieder2020resource, rizun2014simulation}

\subsubsection{Principles and Definitions}

\paragraph{Bias}
Humans demonstrate multiple forms and sources of biases. In one case, humans can demonstrate skewed interpretations of data, probabilities, confidence, etc. \cite{lieder2020resource}. This kind of irrationality and bias is often attributed to the use of cognitive heuristics. It is argued that the use of heuristics causes humans to often accept coarse analysis and sub-optimal solutions. This demonstrates how humans can violate the traditional notions of rational behavior. Further, the use of heuristics and other shortcuts can lead to skewed interpretations of information. These biases and coarse representations can lead to sensitivity regarding the interpretation of rate events, risk, probability, and more. Related, it has also been shown how humans tend to under-react to probabilistic information and also fail to follow belief updates modeled by Bayes' rule \cite{dasgupta2020algorithmic}.

Another aspect of bias can come in the form of inattentiveness, which could be linked to the bounded computational power of human cognition. Humans also demonstrate a resistance to changes in views or opinions when facing contradictory information \cite{mandelbaum2020can}. In fact, it is common for people to become more convicted in their views rather than convinced they might have been wrong. Bias can also come in the form of inductive bias introduced in architecture design and algorithmic choices made by the humans generating algorithms. Both implicitly and explicitly, humans introduce inductive biases into the artificial systems they are developing. In fact, \cite{goyal2020inductive} notes how much of the success of Deep Learning models could be attributed to the inclusion of inductive bias in these systems. Further, the authors note how the use of inductive biases might be a requirement for the creation of generalized artificial intelligence. They note how the biases allow for assumptions regarding the problem being solved or the interpretation of information, which makes the system better suited for adaptation to broader datasets.

\paragraph{Probability-Based Models of Behavior and Reasoning}

To model deficiencies or biases in understanding and reasoning in humans, researchers have investigated Bayesian representations and similar. For instance, (Hierarchical) Bayesian models can demonstrate how humans reason about likelihoods of outcomes and likelihoods of scenarios \cite{tenenbaum2011grow, dasgupta2020algorithmic, dasgupta2020theory, rastogi2020deciding, zhu2020bayesian, ullman2020bayesian}, and how human reasoning shows flawed interpretations or skewed scales of importance (high or low). This illustrates how humans can under-react to probabilities \cite{dasgupta2020algorithmic, dasgupta2020theory} or estimate likelihoods based on observed frequencies. Studies have shown how people tend to select options at a similar frequency to their probability of good outcome rather than developing a bias to the choice with highest likelihood of good outcome \cite{zhu2020bayesian}. Similarly, Bayesian models can replicate the performance of human study participants when tasked with selecting in a multi-arm bandit problem \cite{reverdy2014modeling}. This type of representation of understanding can also be extended to a causal model, which models the cause and effect relationship between different environment aspects \cite{griffiths2005structure}. Causal models are an important aspect of human reasoning as they allow for predictions and retrospective reasoning.

\paragraph{Fairness}
With the use of algorithms, a consideration needs to be made with respect to the fairness of outcomes. As noted in \cite{lee2018understanding}, fairness requires equal or equitable treatment of everyone based on their performance or needs. As a concept, this is straightforward. There should be as little (preferably none) bias in the treatment of individuals. We would prefer the outcome of a human's decision be fair, so it is natural to desire the same behavior from an algorithm. Fairness can both be a quantity measured with respect to a metric \cite{pmlr-v97-heidari19a}, or fairness can represent a qualitative impression people have about a feature or outcome \cite{grgic2018human, lee2018understanding}. These measures provide an indication of how people perceive the features of an algorithm or its behavior. The notion of fairness can also be considered with respect to longer timescales and deeper consequences. In \cite{pmlr-v97-heidari19a}, the longer effects of fairness measures and decisions are weighed to indicate how strongly an approach may impact the behavior of the humans affected.

\subsubsection{Applications and Recent Results}

In \cite{pmlr-v97-heidari19a} the authors model population-level changes at the macro scale that are caused by algorithmic decisions and how this relates to fairness. Using measures of segregation from sociology and economics, they quantified these changes. This allowed them to measure different directions of the shift in the group-conditional distribution based on the different models demonstrated. In this context, the authors note how most notions or measures of fairness assume a static population. They argue that this approach fails to account for long-term welfare and prosperity. Therefore, they propose a measure which considers a notion of \emph{effort} based on economic literature on Equality of Opportunity. Their effort function highlights idea that the necessary changes for a desirable outcome often are more difficult for a member of a disadvantaged group compared to an advantage counterpart. Based on this concept, they formulate \emph{effort unfairness} as the discrepancy in effort required for members in different groups to obtain desired outcomes.

In their approach, they assume individuals imitate an exemplar individual who demonstrates a more desirable algorithm outcome, which is related to social learning. The assumption being that the observer would believe this imitation offers a higher likelihood of a better outcome, which suggests an individual would exert effort to attain a replication of the exemplar's social model if doing so could result in higher overall utility. To model this dynamic, the authors propose a group-dependent, data-driven measure of effort, which is inspired by literature on Equality of Opportunity (EOP). This is noted as the effort it takes individual $i$ to improve their feature $k$ value from $x$ to $x'$. This effort is assumed proportional to the difference between the rank/quantile of $x'$ and $x$ in the distribution of feature $k$ in $i$’s social group. This means an individual which successfully replicates the role model obtains a positive utility (reward minus effort).

In this context, the authors consider a standard supervised learning setting with training data set $D = \{(x_i , y_i)\}^n_{i=1}$ of $n$ instances $z_i = (x_i , y_i)$, where $x_i \in \mathcal{X}$ specifies the feature vector for individual $i$ and $y_i \in\mathcal{Y}$, the ground truth label for him/her. Further, let $s_i$ refer to the sensitive feature value (e.g. race, gender, or their intersection) for individual $i$. To measure the impact of a change in label, the authors define benefit function $b: \mathcal{X} , \mathcal{Y} \times \mathcal{Y} \rightarrow \mathbb{R}$ that quantifies the benefit of a change in label from $y$ to $\hat{y}$ and is assumed a linear function.

To formally define measures of effort, reward, and utility, the authors provide the following. For deployed model $h$ and individual characterized by $z = (x, y)$, let $\mathcal{R}_h(z, z')$ specify the reward/benefit as a result of characteristic change from $z$ to $z'$:
\begin{equation}
    \mathcal{R}_h(z, z') = b(h(x'), y')^\alpha - b(h(x), y)^\alpha,
\end{equation}
where $\alpha > 0$ is a constant specifying the individual’s degree of risk aversion. Again, this demonstrates a measure which considers the impact on an individuals outcome when they change a characteristic. For instance, a person could attain a degree or find a new job. Let $\mathcal{E}_h(z, z')$ specify the effort required for a qualification change from $z$ to $z'$. The overall utility of the individual is then $\mathcal{U}_h(z, z') = \mathcal{R}_h(z, z') - \mathcal{E}_h(z, z')$ (i.e. utility equals reward minus effort).

Based on the above assumptions and definitions, the authors define several effort-based measures of unfairness, including:
\begin{definition}[Effort-reward Unfairness]
    For a predictive model h, the effort-reward unfairness is the inequality of the following metric across different groups:
    \begin{equation}
        \mathbb{E}_{q\sim P:s_i=s}\max_{z\in\mathcal{Z}}\mathcal{U}_h(z_i,z)
    \end{equation}
\end{definition}
This provides a measure of the performance of the highest utility members of a group achieved by a higher level of effort. The authors note the impact of both threshold $\delta$ and the models. It is demonstrated that $\delta$ and the model can cause a significant change in outcome and effort level for the same data.

To measure performance, the authors utilize several techniques from sociology measuring segregation. First, \textbf{evenness} measures how unevenly the minority group is distributed over areal units. The evenness value is maximized when all units have equivalent relative numbers of minority and majority members as the whole population. For a formal definition, the authors use the Atkinson Index (AI), which measures inequality. The second measure, \textbf{centralization} is the degree to which a group is spatially located near the center of an urban area. This can be measured by comparing the percentage of minority residents living in the central areas of the city. The authors utilize the Centralization Index (CI), which is defined as:
\begin{equation}
    \frac{\sum_{i\,\textrm{central}} m_i}{m},
\end{equation}
where $m$ is the total minority population. The third measure of segregation, the Absolute Clustering Index (ACI), ``expresses the average number of [minority] members in nearby [areal units] as a proportion of the total population in those nearby [areal units]".

To demonstrate the impact of differing measures and concepts of fairness, the authors focus
on the case of linear regression. The model was trained by minimizing the mean squared error while imposing welfare constraints defined in \cite{heidari2018fairness}. The results demonstrate the impact of the fairness constraints on the three segregation measures defined above when females are the minority group considered. The authors note that the expectation would be such that the constraints would result in reduced segregation in the long run, but the results demonstrate deviations. Let $\tau$ denotes a minimum threshold on $\mathcal{U}$ (i.e. $\mathcal{U} \geq \tau$) with respect to learning with the loss function $L_D$. The results show that in the case of small $\tau$ values, enforcing fairness constraints can generate a reduction in the degree of clustering. Conversely, larger values of $\tau$ can generate the opposite effect. Regarding evenness, this measure is relatively unchanged for all tested values of $\tau$. The authors note that these results indicate: 1) a label of \emph{desirable} for some members of the disadvantaged group motivates those members to remain unchanged; 2) the positively labeled members can serve as role models for other group members and motivate positive changes in others.

\paragraph{Additional Relevant Results}

As noted above, biases can be represented in both interpretations of information as well as in the exhibited behavior. This can be demonstrated by Bayesian models of cognitive biases \cite{rastogi2020deciding}, biased probability judgments \cite{zhu2020bayesian}, sensitivity to risk/uncertainty \cite{enkhtaivan2021competition}, and similar topics relating to representation and interpretation of data/statistics \cite{dasgupta2020theory, lieder2018overrepresentation}. Biases can also be exhibited in how people respond to others or digital avatars (e.g. racial preferences in representation or interpretation of faces \cite{james2017impact, liao2020racial}). These biases can also influence behavior due to or via heuristics \cite{karlan2021reasoning, lau2001advantages} both in terms of reasoning as well as decision-making (e.g. experience level impacting performance of heuristic outcome). Another factor relating to bias is its impact on model designs and inductive biases introduced into systems \cite{goyal2020inductive, qiu2020learning, lakkaraju2017selective}. These biases can impact how a system is formed, how it utilizes contextual information, or how the system might need to compensate for unobserved factors available at decision time when learning from past behaviors of humans. Alternatively, it can also be important to avoid introducing biases caused by algorithms making determinations or learning from data which might be corrupted. The approaches in \cite{thulasidasan2019combating} and \cite{ramaswamy2018consistent} demonstrate an effort to enable systems with the ability signal when they are uncertain whether a label for an input item is accurate via an abstention action. This can make systems more accurate as well as less susceptible to perturbation-based attacks or label noise. An additional context for bias is the aspect of fairness \cite{backurs2019scalable, asudeh2019designing, kannan2019downstream, corbett2017algorithmic}. This can be in relation to both the representation provided to the algorithm as well the decision or outcome. Systems which exhibit unfair biases can generate disadvantages when the outcome has impacts on decisions relating to humans. For instance, non-uniformity or under-representation of samples in data can bias the performance of the trained model. If the lack of representation is in relation to race or gender, this can cause sociological consequences.

\section{Conclusion}
\label{sec:conclusions}

This paper completes the survey on approaches to model the human behavior in Human-Centric AI systems, complementing areas covered in~\cite{fuchs2022partI}. In general, modeling the human behavior is seen as a cornerstone of future autonomous and adaptive systems, which can unleash full integration between humans and AI agents working in tandem. It will allow AI agents to make their choices more comprehensible by users, on the one hand, and AI agents to interpret, predict and take account of human actions and choices, on the other hand. While in~\cite{fuchs2022partI} we have covered approaches where AI agents learn human behavior by trial-and-error, as well as approaches ``coding" directly models of human reasoning, in this paper we have focused on approaches taking into account limitation of cognitive resources and how to optimize the, as well as approaches modeling uncertainty in reasoning and irrational behaviors.

In Section~\ref{sec:bounded_rationality_and_cog_limits} (\emph{Bounded Rationality and Cognitive Limitations}), we outlined and discussed approaches which can or attempt to replicate how humans gain/utilize knowledge to perform tasks, taking in particular account an efficient use of (limited) cognitive resources. In these cases it is often necessary to provide models of world dynamics and actions available so the model can replicate human performance, which clearly can offer effective performance. However, the models required to generate solutions typically require domain knowledge and so can be difficult or costly to generate. For instance, architectures like those described in Section~\ref{sec:cog_architectures}, typically rely on models of world and cognitive dynamics to enable reasoning over available capabilities given the bounds on time and cognitive cost. Still, these approaches can be effective when trying to better align with how humans perform tasks and allocate cognitive resources. On the other hand, these approaches can also be sample inefficient or require detailed (possibly expensive) models of knowledge. Related, behavior and reasoning can be represented through the use of heuristics mimicking the two systems humans are believed to utilize as seen in Section~\ref{sec:cog_heuristics}. The use of these heuristics has similar benefits and pitfalls. Again, these models often prove effective and are efficient to utilize, but often rely on domain knowledge or potentially coarse representations of human reasoning. The representation of reasoning with heuristics demonstrates an approach to model the assumed cognitive bounds on human reasoning. Such boundedness is also related to the assumption of bounded rationality found in Game Theory (Section~\ref{sec:game_theory}). This assumption allows for constraints on the representation which enables generation of solutions. These solutions optimize behavior based on these assumptions as they enable reasoning over the behavior of all players in a system based on their expected payoffs for possible outcomes. This representation enables strong guarantees and methods, but can often be conservative and a less than accurate portrayal of actual human behavior (e.g. players may not behave rationally). With another focus, in order to integrate capabilities related to reasoning and how humans utilize stimuli given their available cognitive resources, researchers have investigated methods for mimicking how humans attend to information. In these approaches, as discussed in Section~\ref{sec:attention}, systems learn a model for masking and associating input values to learn a deeper understanding of the stimulus. These models have shown strong performance in language and behavior scenarios and can allow for more abstract representations supporting more variation in the scenario composition. On the other hand, as with many Deep Learning scenarios, these methods can be difficult to train or require a large amount of samples, so careful consideration is needed when defining an approach.

For the remainder of the paper, in Section~\ref{sec:uncertainty_and_irrationality} (\emph{Uncertainty and Irrationality}), we discuss approaches which represent methods reasoning or considering irrationality, biases, or beliefs of humans. In this context, the presented approaches learn or use models of belief. This can be the effect of beliefs with respect to how it introduces biases in behavior and reasoning (see Section~\ref{sec:bias_and_fairness}). Similar to the previous topic, this also requires development of models, which can be costly and challenging. On the other hand, these methods can offer effective means for measuring the impact of biases in behavior or reasoning with respect to future outcomes or fairness. In addition to biases, the process of human reasoning involves moments of uncertainty. In this case, the decision-maker exists in a state of uncertainty as they transition from an initial state to a decision. These transitions represent new knowledge or changes in belief, which impact the final outcome. We discussed one approach for representing uncertainty in reasoning in Section~\ref{sec:quantum_representation}. Such a representation utilizes the flexibility of quantum mechanisms and the aspect of the interference effect for unobserved paths rather than the assumption of a single path seen in Markovian models. This improves the flexibility of representation and allows for representations which model fallacies in reasoning (e.g. disjunction fallacy), but also increases the complexity of the resulting models.

All in all, we can conclude there is not a one-size-fits-all approach to best model the human behavior in HCAI. Each approach has its onw pros and flip side, and the best choice largely depends on the specific problem at hand. However, the literature on this topic is very vast and articulated, and this allows designers of HCAI systems to leverage an extensive toolbox to equip AI agents with practical approaches to model human behavior. Therefore, we believe this area is going to emerge as one of the most active in the coming years, in the field cutting across autonomous and adaptive systems, pervasive environments, and advanced AI agents.

\section{Acknowledgments}\label{sec:acknowledgments}

This work was supported by the H2020 Humane-AI-Net project (grant \#952026) and by the CHIST-ERA grant CHIST-ERA-19-XAI-010, by MUR (grant No. not yet available), FWF (grant No. I 5205), EPSRC (grant No. EP/V055712/1), NCN (grant No. 2020/02/Y/ST6/00064), ETAg (grant No. SLTAT21096), BNSF (grant No. \rn{KP}-06-\rn{DOO}2/5).

\bibliographystyle{ACM-Reference-Format}
\bibliography{main}


\begin{thebibliography}{139}

%%% the Web address.

\ifx \showCODEN    \undefined \def \showCODEN     #1{\unskip}     \fi
\ifx \showDOI      \undefined \def \showDOI       #1{#1}\fi
\ifx \showISBNx    \undefined \def \showISBNx     #1{\unskip}     \fi
\ifx \showISBNxiii \undefined \def \showISBNxiii  #1{\unskip}     \fi
\ifx \showISSN     \undefined \def \showISSN      #1{\unskip}     \fi
\ifx \showLCCN     \undefined \def \showLCCN      #1{\unskip}     \fi
\ifx \shownote     \undefined \def \shownote      #1{#1}          \fi
\ifx \showarticletitle \undefined \def \showarticletitle #1{#1}   \fi
\ifx \showURL      \undefined \def \showURL       {\relax}        \fi
% invisible to TeX
\providecommand\bibfield[2]{#2}
\providecommand\bibinfo[2]{#2}
\providecommand\natexlab[1]{#1}
\providecommand\showeprint[2][]{arXiv:#2}

\bibitem[Aerts et~al\mbox{.}(2021)]%
        {aerts2021modeling}
\bibfield{author}{\bibinfo{person}{Diederik Aerts},
  \bibinfo{person}{Massimiliano~Sassoli De~Bianchi}, \bibinfo{person}{Sandro
  Sozzo}, {and} \bibinfo{person}{Tomas Veloz}.}
  \bibinfo{year}{2021}\natexlab{}.
\newblock \showarticletitle{Modeling human decision-making: An overview of the
  Brussels quantum approach}.
\newblock \bibinfo{journal}{\emph{Foundations of Science}}
  \bibinfo{volume}{26}, \bibinfo{number}{1} (\bibinfo{year}{2021}),
  \bibinfo{pages}{27--54}.
\newblock


\bibitem[Anderson(2013)]%
        {anderson2013adaptive}
\bibfield{author}{\bibinfo{person}{John~R Anderson}.}
  \bibinfo{year}{2013}\natexlab{}.
\newblock \bibinfo{booktitle}{\emph{The adaptive character of thought}}.
\newblock \bibinfo{publisher}{Psychology Press}.
\newblock


\bibitem[Anderson et~al\mbox{.}(1996)]%
        {anderson1996working}
\bibfield{author}{\bibinfo{person}{John~R Anderson}, \bibinfo{person}{Lynne~M
  Reder}, {and} \bibinfo{person}{Christian Lebiere}.}
  \bibinfo{year}{1996}\natexlab{}.
\newblock \showarticletitle{Working memory: Activation limitations on
  retrieval}.
\newblock \bibinfo{journal}{\emph{Cognitive psychology}} \bibinfo{volume}{30},
  \bibinfo{number}{3} (\bibinfo{year}{1996}), \bibinfo{pages}{221--256}.
\newblock


\bibitem[Arnaboldi et~al\mbox{.}(2017)]%
        {arnaboldi2017online}
\bibfield{author}{\bibinfo{person}{Valerio Arnaboldi}, \bibinfo{person}{Marco
  Conti}, \bibinfo{person}{Andrea Passarella}, {and} \bibinfo{person}{Robin~IM
  Dunbar}.} \bibinfo{year}{2017}\natexlab{}.
\newblock \showarticletitle{Online social networks and information diffusion:
  The role of ego networks}.
\newblock \bibinfo{journal}{\emph{Online Social Networks and Media}}
  \bibinfo{volume}{1} (\bibinfo{year}{2017}), \bibinfo{pages}{44--55}.
\newblock


\bibitem[Asgari and Beauregard(2021)]%
        {asgaribrain}
\bibfield{author}{\bibinfo{person}{Alireza Asgari} {and} \bibinfo{person}{Yvan
  Beauregard}.} \bibinfo{year}{2021}\natexlab{}.
\newblock \showarticletitle{Brain-Inspired Model for Decision-Making in the
  Selection of Beneficial Information Among Signals Received by an
  Unpredictable Information-Development Environment}.
\newblock \bibinfo{journal}{\emph{angrxiv}} (\bibinfo{year}{2021}).
\newblock
\urldef\tempurl%
\url{https://engrxiv.org/preprint/view/1525}
\showURL{%
\tempurl}


\bibitem[Askari et~al\mbox{.}(2019)]%
        {askari2019behavioral}
\bibfield{author}{\bibinfo{person}{Gholamreza Askari},
  \bibinfo{person}{Madjid~Eshaghi Gordji}, {and} \bibinfo{person}{Choonkil
  Park}.} \bibinfo{year}{2019}\natexlab{}.
\newblock \showarticletitle{The behavioral model and game theory}.
\newblock \bibinfo{journal}{\emph{Palgrave Communications}}
  \bibinfo{volume}{5}, \bibinfo{number}{1} (\bibinfo{year}{2019}),
  \bibinfo{pages}{1--8}.
\newblock


\bibitem[Asudeh et~al\mbox{.}(2019)]%
        {asudeh2019designing}
\bibfield{author}{\bibinfo{person}{Abolfazl Asudeh}, \bibinfo{person}{HV
  Jagadish}, \bibinfo{person}{Julia Stoyanovich}, {and} \bibinfo{person}{Gautam
  Das}.} \bibinfo{year}{2019}\natexlab{}.
\newblock \showarticletitle{Designing fair ranking schemes}. In
  \bibinfo{booktitle}{\emph{Proceedings of the 2019 International Conference on
  Management of Data}}. \bibinfo{pages}{1259--1276}.
\newblock


\bibitem[Backurs et~al\mbox{.}(2019)]%
        {backurs2019scalable}
\bibfield{author}{\bibinfo{person}{Arturs Backurs}, \bibinfo{person}{Piotr
  Indyk}, \bibinfo{person}{Krzysztof Onak}, \bibinfo{person}{Baruch Schieber},
  \bibinfo{person}{Ali Vakilian}, {and} \bibinfo{person}{Tal Wagner}.}
  \bibinfo{year}{2019}\natexlab{}.
\newblock \showarticletitle{Scalable fair clustering}. In
  \bibinfo{booktitle}{\emph{International Conference on Machine Learning}}.
  PMLR, \bibinfo{pages}{405--413}.
\newblock


\bibitem[Booch et~al\mbox{.}(2020)]%
        {booch2020thinking}
\bibfield{author}{\bibinfo{person}{Grady Booch}, \bibinfo{person}{Francesco
  Fabiano}, \bibinfo{person}{Lior Horesh}, \bibinfo{person}{Kiran Kate},
  \bibinfo{person}{Jon Lenchner}, \bibinfo{person}{Nick Linck},
  \bibinfo{person}{Andrea Loreggia}, \bibinfo{person}{Keerthiram Murugesan},
  \bibinfo{person}{Nicholas Mattei}, \bibinfo{person}{Francesca Rossi},
  {et~al\mbox{.}}} \bibinfo{year}{2020}\natexlab{}.
\newblock \showarticletitle{Thinking fast and slow in ai}.
\newblock \bibinfo{journal}{\emph{arXiv preprint arXiv:2010.06002}}
  (\bibinfo{year}{2020}).
\newblock


\bibitem[Busemeyer et~al\mbox{.}(2006)]%
        {busemeyer2006quantum}
\bibfield{author}{\bibinfo{person}{Jerome~R Busemeyer}, \bibinfo{person}{Zheng
  Wang}, {and} \bibinfo{person}{James~T Townsend}.}
  \bibinfo{year}{2006}\natexlab{}.
\newblock \showarticletitle{Quantum dynamics of human decision-making}.
\newblock \bibinfo{journal}{\emph{Journal of Mathematical Psychology}}
  \bibinfo{volume}{50}, \bibinfo{number}{3} (\bibinfo{year}{2006}),
  \bibinfo{pages}{220--241}.
\newblock


\bibitem[Conti et~al\mbox{.}(2013)]%
        {conti2013design}
\bibfield{author}{\bibinfo{person}{Marco Conti}, \bibinfo{person}{Matteo
  Mordacchini}, {and} \bibinfo{person}{Andrea Passarella}.}
  \bibinfo{year}{2013}\natexlab{}.
\newblock \showarticletitle{Design and performance evaluation of data
  dissemination systems for opportunistic networks based on cognitive
  heuristics}.
\newblock \bibinfo{journal}{\emph{ACM Transactions on Autonomous and Adaptive
  Systems (TAAS)}} \bibinfo{volume}{8}, \bibinfo{number}{3}
  (\bibinfo{year}{2013}), \bibinfo{pages}{1--32}.
\newblock


\bibitem[Conti et~al\mbox{.}(2017)]%
        {CONTI20171}
\bibfield{author}{\bibinfo{person}{Marco Conti}, \bibinfo{person}{Andrea
  Passarella}, {and} \bibinfo{person}{Sajal~K. Das}.}
  \bibinfo{year}{2017}\natexlab{}.
\newblock \showarticletitle{The Internet of People (IoP): A new wave in
  pervasive mobile computing}.
\newblock \bibinfo{journal}{\emph{Pervasive and Mobile Computing}}
  \bibinfo{volume}{41} (\bibinfo{year}{2017}), \bibinfo{pages}{1--27}.
\newblock
\urldef\tempurl%
\url{https://doi.org/10.1016/j.pmcj.2017.07.009}
\showDOI{\tempurl}


\bibitem[Corbett-Davies et~al\mbox{.}(2017)]%
        {corbett2017algorithmic}
\bibfield{author}{\bibinfo{person}{Sam Corbett-Davies}, \bibinfo{person}{Emma
  Pierson}, \bibinfo{person}{Avi Feller}, \bibinfo{person}{Sharad Goel}, {and}
  \bibinfo{person}{Aziz Huq}.} \bibinfo{year}{2017}\natexlab{}.
\newblock \showarticletitle{Algorithmic decision making and the cost of
  fairness}. In \bibinfo{booktitle}{\emph{Proceedings of the 23rd acm sigkdd
  international conference on knowledge discovery and data mining}}.
  \bibinfo{pages}{797--806}.
\newblock


\bibitem[Correia and Colombini(2021)]%
        {correia2021attention}
\bibfield{author}{\bibinfo{person}{Alana de~Santana Correia} {and}
  \bibinfo{person}{Esther~Luna Colombini}.} \bibinfo{year}{2021}\natexlab{}.
\newblock \showarticletitle{Attention, please! A survey of Neural Attention
  Models in Deep Learning}.
\newblock \bibinfo{journal}{\emph{arXiv preprint arXiv:2103.16775}}
  (\bibinfo{year}{2021}).
\newblock


\bibitem[Cranford et~al\mbox{.}(2020a)]%
        {cranford2020adaptive}
\bibfield{author}{\bibinfo{person}{Edward Cranford}, \bibinfo{person}{Cleotilde
  Gonzalez}, \bibinfo{person}{Palvi Aggarwal}, \bibinfo{person}{Sarah Cooney},
  \bibinfo{person}{Milind Tambe}, {and} \bibinfo{person}{Christian Lebiere}.}
  \bibinfo{year}{2020}\natexlab{a}.
\newblock \showarticletitle{Adaptive cyber deception: Cognitively informed
  signaling for cyber defense}. In \bibinfo{booktitle}{\emph{Proceedings of the
  53rd Hawaii International Conference on System Sciences}}.
\newblock


\bibitem[Cranford et~al\mbox{.}(2020b)]%
        {cranford2020toward}
\bibfield{author}{\bibinfo{person}{Edward~A Cranford},
  \bibinfo{person}{Cleotilde Gonzalez}, \bibinfo{person}{Palvi Aggarwal},
  \bibinfo{person}{Sarah Cooney}, \bibinfo{person}{Milind Tambe}, {and}
  \bibinfo{person}{Christian Lebiere}.} \bibinfo{year}{2020}\natexlab{b}.
\newblock \showarticletitle{Toward Personalized Deceptive Signaling for Cyber
  Defense Using Cognitive Models}.
\newblock \bibinfo{journal}{\emph{Topics in Cognitive Science}}
  \bibinfo{volume}{12}, \bibinfo{number}{3} (\bibinfo{year}{2020}),
  \bibinfo{pages}{992--1011}.
\newblock


\bibitem[Cranford et~al\mbox{.}(2020c)]%
        {cranford2020cognitive}
\bibfield{author}{\bibinfo{person}{Edward~A Cranford},
  \bibinfo{person}{Sterling Somers}, \bibinfo{person}{Konstantinos
  Mitsopoulos}, {and} \bibinfo{person}{Christian Lebiere}.}
  \bibinfo{year}{2020}\natexlab{c}.
\newblock \showarticletitle{Cognitive salience of features in cyber-attacker
  decision making}. In \bibinfo{booktitle}{\emph{Proceedings of the 18th annual
  meeting of the international conference on cognitive modeling. University
  Park, PA: Applied Cognitive Science Lab, Penn State}}.
\newblock


\bibitem[Dasgupta(2020)]%
        {dasgupta2020algorithmic}
\bibfield{author}{\bibinfo{person}{Ishita Dasgupta}.}
  \bibinfo{year}{2020}\natexlab{}.
\newblock \emph{\bibinfo{title}{Algorithmic approaches to ecological
  rationality in humans and machines}}.
\newblock \bibinfo{thesistype}{Ph.\,D. Dissertation}. \bibinfo{school}{Harvard
  University}.
\newblock


\bibitem[Dasgupta et~al\mbox{.}(2020)]%
        {dasgupta2020theory}
\bibfield{author}{\bibinfo{person}{Ishita Dasgupta}, \bibinfo{person}{Eric
  Schulz}, \bibinfo{person}{Joshua~B Tenenbaum}, {and}
  \bibinfo{person}{Samuel~J Gershman}.} \bibinfo{year}{2020}\natexlab{}.
\newblock \showarticletitle{A theory of learning to infer.}
\newblock \bibinfo{journal}{\emph{Psychological review}} \bibinfo{volume}{127},
  \bibinfo{number}{3} (\bibinfo{year}{2020}), \bibinfo{pages}{412}.
\newblock


\bibitem[Dehdashti et~al\mbox{.}(2020)]%
        {dehdashti2020irrationality}
\bibfield{author}{\bibinfo{person}{Shahram Dehdashti}, \bibinfo{person}{Lauren
  Fell}, {and} \bibinfo{person}{Peter Bruza}.} \bibinfo{year}{2020}\natexlab{}.
\newblock \showarticletitle{On the irrationality of being in two minds}.
\newblock \bibinfo{journal}{\emph{Entropy}} \bibinfo{volume}{22},
  \bibinfo{number}{2} (\bibinfo{year}{2020}), \bibinfo{pages}{174}.
\newblock


\bibitem[Dimov et~al\mbox{.}(2020)]%
        {dimov2020model}
\bibfield{author}{\bibinfo{person}{Cvetomir Dimov}, \bibinfo{person}{Patrick~H
  Khader}, \bibinfo{person}{Julian~N Marewski}, {and} \bibinfo{person}{Thorsten
  Pachur}.} \bibinfo{year}{2020}\natexlab{}.
\newblock \showarticletitle{How to model the neurocognitive dynamics of
  decision making: A methodological primer with ACT-R}.
\newblock \bibinfo{journal}{\emph{Behavior research methods}}
  \bibinfo{volume}{52}, \bibinfo{number}{2} (\bibinfo{year}{2020}),
  \bibinfo{pages}{857--880}.
\newblock


\bibitem[Dobson et~al\mbox{.}(2019)]%
        {dobson2019integrating}
\bibfield{author}{\bibinfo{person}{Andrew~DM Dobson}, \bibinfo{person}{Emiel
  De~Lange}, \bibinfo{person}{Aidan Keane}, \bibinfo{person}{Harriet Ibbett},
  {and} \bibinfo{person}{EJ Milner-Gulland}.} \bibinfo{year}{2019}\natexlab{}.
\newblock \showarticletitle{Integrating models of human behaviour between the
  individual and population levels to inform conservation interventions}.
\newblock \bibinfo{journal}{\emph{Philosophical Transactions of the Royal
  Society B}} \bibinfo{volume}{374}, \bibinfo{number}{1781}
  (\bibinfo{year}{2019}), \bibinfo{pages}{20180053}.
\newblock


\bibitem[Dolfin et~al\mbox{.}(2017)]%
        {dolfin2017modeling}
\bibfield{author}{\bibinfo{person}{M Dolfin}, \bibinfo{person}{L Leonida},
  {and} \bibinfo{person}{N Outada}.} \bibinfo{year}{2017}\natexlab{}.
\newblock \showarticletitle{Modeling human behavior in economics and social
  science}.
\newblock \bibinfo{journal}{\emph{Physics of life reviews}}
  \bibinfo{volume}{22} (\bibinfo{year}{2017}), \bibinfo{pages}{1--21}.
\newblock


\bibitem[Driessens and Ramon(2003)]%
        {driessens2003relational}
\bibfield{author}{\bibinfo{person}{Kurt Driessens} {and} \bibinfo{person}{Jan
  Ramon}.} \bibinfo{year}{2003}\natexlab{}.
\newblock \showarticletitle{Relational instance based regression for relational
  reinforcement learning}. In \bibinfo{booktitle}{\emph{Proceedings of the 20th
  International Conference on Machine Learning (ICML-03)}}.
  \bibinfo{pages}{123--130}.
\newblock


\bibitem[Eiband et~al\mbox{.}(2021)]%
        {eiband2021support}
\bibfield{author}{\bibinfo{person}{Malin Eiband}, \bibinfo{person}{Daniel
  Buschek}, {and} \bibinfo{person}{Heinrich Hussmann}.}
  \bibinfo{year}{2021}\natexlab{}.
\newblock \showarticletitle{How to support users in understanding intelligent
  systems? Structuring the discussion}. In \bibinfo{booktitle}{\emph{26th
  International Conference on Intelligent User Interfaces}}.
  \bibinfo{pages}{120--132}.
\newblock


\bibitem[Enkhtaivan et~al\mbox{.}(2021)]%
        {enkhtaivan2021competition}
\bibfield{author}{\bibinfo{person}{Enkhzaya Enkhtaivan}, \bibinfo{person}{Joel
  Nishimura}, \bibinfo{person}{Cheng Ly}, {and} \bibinfo{person}{Amy~L
  Cochran}.} \bibinfo{year}{2021}\natexlab{}.
\newblock \showarticletitle{A competition of critics in human decision-making}.
\newblock \bibinfo{journal}{\emph{Computational Psychiatry}}
  \bibinfo{volume}{5}, \bibinfo{number}{1} (\bibinfo{year}{2021}).
\newblock


\bibitem[Facione and Gittens(2012)]%
        {facione2012think}
\bibfield{author}{\bibinfo{person}{P.A. Facione} {and} \bibinfo{person}{C.A.
  Gittens}.} \bibinfo{year}{2012}\natexlab{}.
\newblock \bibinfo{booktitle}{\emph{Think Critically}}.
\newblock \bibinfo{publisher}{Pearson}.
\newblock
\showISBNx{9780205490981}
\showLCCN{2011050397}
\urldef\tempurl%
\url{https://books.google.it/books?id=YGM5ygAACAAJ}
\showURL{%
\tempurl}


\bibitem[Fernando et~al\mbox{.}(2020)]%
        {fernando2020deep}
\bibfield{author}{\bibinfo{person}{Tharindu Fernando}, \bibinfo{person}{Simon
  Denman}, \bibinfo{person}{Sridha Sridharan}, {and} \bibinfo{person}{Clinton
  Fookes}.} \bibinfo{year}{2020}\natexlab{}.
\newblock \showarticletitle{Deep inverse reinforcement learning for behavior
  prediction in autonomous driving: Accurate forecasts of vehicle motion}.
\newblock \bibinfo{journal}{\emph{IEEE Signal Processing Magazine}}
  \bibinfo{volume}{38}, \bibinfo{number}{1} (\bibinfo{year}{2020}),
  \bibinfo{pages}{87--96}.
\newblock


\bibitem[Ferret et~al\mbox{.}(2019)]%
        {ferret2019self}
\bibfield{author}{\bibinfo{person}{Johan Ferret}, \bibinfo{person}{Rapha{\"e}l
  Marinier}, \bibinfo{person}{Matthieu Geist}, {and} \bibinfo{person}{Olivier
  Pietquin}.} \bibinfo{year}{2019}\natexlab{}.
\newblock \showarticletitle{Self-attentional credit assignment for transfer in
  reinforcement learning}.
\newblock \bibinfo{journal}{\emph{arXiv preprint arXiv:1907.08027}}
  (\bibinfo{year}{2019}).
\newblock


\bibitem[Fuchs et~al\mbox{.}(2022)]%
        {fuchs2022partI}
\bibfield{author}{\bibinfo{person}{Andrew Fuchs}, \bibinfo{person}{Andrea
  Passarella}, {and} \bibinfo{person}{Marco Conti}.}
  \bibinfo{year}{2022}\natexlab{}.
\newblock \showarticletitle{Modeling Human Behavior Part I - Cognitive
  approaches and Uncertainty}.
\newblock \bibinfo{journal}{\emph{arXiv preprint arXiv:XXX}}
  (\bibinfo{year}{2022}).
\newblock


\bibitem[Gigerenzer(2008)]%
        {gigerenzer2008moral}
\bibfield{author}{\bibinfo{person}{Gerd Gigerenzer}.}
  \bibinfo{year}{2008}\natexlab{}.
\newblock \showarticletitle{Moral intuition= fast and frugal heuristics?}
\newblock In \bibinfo{booktitle}{\emph{Moral psychology}}.
  \bibinfo{publisher}{MIT Press}, \bibinfo{pages}{1--26}.
\newblock


\bibitem[Gonzalez et~al\mbox{.}(2003)]%
        {gonzalez2003instance}
\bibfield{author}{\bibinfo{person}{Cleotilde Gonzalez},
  \bibinfo{person}{Javier~F Lerch}, {and} \bibinfo{person}{Christian Lebiere}.}
  \bibinfo{year}{2003}\natexlab{}.
\newblock \showarticletitle{Instance-based learning in dynamic decision
  making}.
\newblock \bibinfo{journal}{\emph{Cognitive Science}} \bibinfo{volume}{27},
  \bibinfo{number}{4} (\bibinfo{year}{2003}), \bibinfo{pages}{591--635}.
\newblock


\bibitem[Goyal and Bengio(2020)]%
        {goyal2020inductive}
\bibfield{author}{\bibinfo{person}{Anirudh Goyal} {and} \bibinfo{person}{Yoshua
  Bengio}.} \bibinfo{year}{2020}\natexlab{}.
\newblock \showarticletitle{Inductive biases for deep learning of higher-level
  cognition}.
\newblock \bibinfo{journal}{\emph{arXiv preprint arXiv:2011.15091}}
  (\bibinfo{year}{2020}).
\newblock


\bibitem[Graziano(2019)]%
        {graziano2019attributing}
\bibfield{author}{\bibinfo{person}{Michael~SA Graziano}.}
  \bibinfo{year}{2019}\natexlab{}.
\newblock \showarticletitle{Attributing awareness to others: the attention
  schema theory and its relationship to behavioural prediction}.
\newblock \bibinfo{journal}{\emph{Journal of Consciousness Studies}}
  \bibinfo{volume}{26}, \bibinfo{number}{3-4} (\bibinfo{year}{2019}),
  \bibinfo{pages}{17--37}.
\newblock


\bibitem[Grgic-Hlaca et~al\mbox{.}(2018)]%
        {grgic2018human}
\bibfield{author}{\bibinfo{person}{Nina Grgic-Hlaca}, \bibinfo{person}{Elissa~M
  Redmiles}, \bibinfo{person}{Krishna~P Gummadi}, {and} \bibinfo{person}{Adrian
  Weller}.} \bibinfo{year}{2018}\natexlab{}.
\newblock \showarticletitle{Human perceptions of fairness in algorithmic
  decision making: A case study of criminal risk prediction}. In
  \bibinfo{booktitle}{\emph{Proceedings of the 2018 World Wide Web
  Conference}}. \bibinfo{pages}{903--912}.
\newblock


\bibitem[Griffiths and Tenenbaum(2005)]%
        {griffiths2005structure}
\bibfield{author}{\bibinfo{person}{Thomas~L Griffiths} {and}
  \bibinfo{person}{Joshua~B Tenenbaum}.} \bibinfo{year}{2005}\natexlab{}.
\newblock \showarticletitle{Structure and strength in causal induction}.
\newblock \bibinfo{journal}{\emph{Cognitive psychology}} \bibinfo{volume}{51},
  \bibinfo{number}{4} (\bibinfo{year}{2005}), \bibinfo{pages}{334--384}.
\newblock


\bibitem[Groeneveld et~al\mbox{.}(2017)]%
        {groeneveld2017theoretical}
\bibfield{author}{\bibinfo{person}{J{\"u}rgen Groeneveld},
  \bibinfo{person}{Birgit M{\"u}ller}, \bibinfo{person}{Carsten~M Buchmann},
  \bibinfo{person}{Gunnar Dressler}, \bibinfo{person}{Cheng Guo},
  \bibinfo{person}{Niklas Hase}, \bibinfo{person}{Falk Hoffmann},
  \bibinfo{person}{F John}, \bibinfo{person}{Christian Klassert},
  \bibinfo{person}{T Lauf}, {et~al\mbox{.}}} \bibinfo{year}{2017}\natexlab{}.
\newblock \showarticletitle{Theoretical foundations of human decision-making in
  agent-based land use models--A review}.
\newblock \bibinfo{journal}{\emph{Environmental modelling \& software}}
  \bibinfo{volume}{87} (\bibinfo{year}{2017}), \bibinfo{pages}{39--48}.
\newblock


\bibitem[Guidotti et~al\mbox{.}(2018)]%
        {10.1145/3236009}
\bibfield{author}{\bibinfo{person}{Riccardo Guidotti}, \bibinfo{person}{Anna
  Monreale}, \bibinfo{person}{Salvatore Ruggieri}, \bibinfo{person}{Franco
  Turini}, \bibinfo{person}{Fosca Giannotti}, {and} \bibinfo{person}{Dino
  Pedreschi}.} \bibinfo{year}{2018}\natexlab{}.
\newblock \showarticletitle{A Survey of Methods for Explaining Black Box
  Models}.
\newblock \bibinfo{journal}{\emph{ACM Comput. Surv.}} \bibinfo{volume}{51},
  \bibinfo{number}{5} (\bibinfo{year}{2018}).
\newblock
\urldef\tempurl%
\url{https://doi.org/10.1145/3236009}
\showDOI{\tempurl}


\bibitem[Gurcan et~al\mbox{.}(2021)]%
        {gurcan2021mapping}
\bibfield{author}{\bibinfo{person}{Fatih Gurcan}, \bibinfo{person}{Nergiz~Ercil
  Cagiltay}, {and} \bibinfo{person}{Kursat Cagiltay}.}
  \bibinfo{year}{2021}\natexlab{}.
\newblock \showarticletitle{Mapping human--computer interaction research themes
  and trends from its existence to today: A topic modeling-based review of past
  60 years}.
\newblock \bibinfo{journal}{\emph{International Journal of Human--Computer
  Interaction}} \bibinfo{volume}{37}, \bibinfo{number}{3}
  (\bibinfo{year}{2021}), \bibinfo{pages}{267--280}.
\newblock


\bibitem[He and Jiang(2018)]%
        {he2018evidential}
\bibfield{author}{\bibinfo{person}{Zichang He} {and} \bibinfo{person}{Wen
  Jiang}.} \bibinfo{year}{2018}\natexlab{}.
\newblock \showarticletitle{An evidential dynamical model to predict the
  interference effect of categorization on decision making results}.
\newblock \bibinfo{journal}{\emph{Knowledge-Based Systems}}
  \bibinfo{volume}{150} (\bibinfo{year}{2018}), \bibinfo{pages}{139--149}.
\newblock


\bibitem[Heidari et~al\mbox{.}(2018)]%
        {heidari2018fairness}
\bibfield{author}{\bibinfo{person}{Hoda Heidari}, \bibinfo{person}{Claudio
  Ferrari}, \bibinfo{person}{Krishna Gummadi}, {and} \bibinfo{person}{Andreas
  Krause}.} \bibinfo{year}{2018}\natexlab{}.
\newblock \showarticletitle{Fairness behind a veil of ignorance: A welfare
  analysis for automated decision making}.
\newblock \bibinfo{journal}{\emph{Advances in Neural Information Processing
  Systems}}  \bibinfo{volume}{31} (\bibinfo{year}{2018}).
\newblock


\bibitem[Heidari et~al\mbox{.}(2019)]%
        {pmlr-v97-heidari19a}
\bibfield{author}{\bibinfo{person}{Hoda Heidari}, \bibinfo{person}{Vedant
  Nanda}, {and} \bibinfo{person}{Krishna Gummadi}.}
  \bibinfo{year}{2019}\natexlab{}.
\newblock \showarticletitle{On the Long-term Impact of Algorithmic Decision
  Policies: Effort Unfairness and Feature Segregation through Social Learning}.
  In \bibinfo{booktitle}{\emph{Proceedings of the 36th International Conference
  on Machine Learning}} \emph{(\bibinfo{series}{Proceedings of Machine Learning
  Research}, Vol.~\bibinfo{volume}{97})},
  \bibfield{editor}{\bibinfo{person}{Kamalika Chaudhuri} {and}
  \bibinfo{person}{Ruslan Salakhutdinov}} (Eds.). \bibinfo{publisher}{PMLR},
  \bibinfo{pages}{2692--2701}.
\newblock
\urldef\tempurl%
\url{https://proceedings.mlr.press/v97/heidari19a.html}
\showURL{%
\tempurl}


\bibitem[Holzinger et~al\mbox{.}(2019)]%
        {holzinger2019interactive}
\bibfield{author}{\bibinfo{person}{Andreas Holzinger}, \bibinfo{person}{Markus
  Plass}, \bibinfo{person}{Michael Kickmeier-Rust}, \bibinfo{person}{Katharina
  Holzinger}, \bibinfo{person}{Gloria~Cerasela Cri{\c{s}}an},
  \bibinfo{person}{Camelia-M Pintea}, {and} \bibinfo{person}{Vasile Palade}.}
  \bibinfo{year}{2019}\natexlab{}.
\newblock \showarticletitle{Interactive machine learning: experimental evidence
  for the human in the algorithmic loop}.
\newblock \bibinfo{journal}{\emph{Applied Intelligence}} \bibinfo{volume}{49},
  \bibinfo{number}{7} (\bibinfo{year}{2019}), \bibinfo{pages}{2401--2414}.
\newblock


\bibitem[Huang et~al\mbox{.}(2019)]%
        {huang2019uncertainty}
\bibfield{author}{\bibinfo{person}{Zhiming Huang}, \bibinfo{person}{Lin Yang},
  {and} \bibinfo{person}{Wen Jiang}.} \bibinfo{year}{2019}\natexlab{}.
\newblock \showarticletitle{Uncertainty measurement with belief entropy on the
  interference effect in the quantum-like Bayesian Networks}.
\newblock \bibinfo{journal}{\emph{Appl. Math. Comput.}}  \bibinfo{volume}{347}
  (\bibinfo{year}{2019}), \bibinfo{pages}{417--428}.
\newblock


\bibitem[Jackson et~al\mbox{.}(2017)]%
        {jackson2017agent}
\bibfield{author}{\bibinfo{person}{Joshua~Conrad Jackson},
  \bibinfo{person}{David Rand}, \bibinfo{person}{Kevin Lewis},
  \bibinfo{person}{Michael~I Norton}, {and} \bibinfo{person}{Kurt Gray}.}
  \bibinfo{year}{2017}\natexlab{}.
\newblock \showarticletitle{Agent-based modeling: A guide for social
  psychologists}.
\newblock \bibinfo{journal}{\emph{Social Psychological and Personality
  Science}} \bibinfo{volume}{8}, \bibinfo{number}{4} (\bibinfo{year}{2017}),
  \bibinfo{pages}{387--395}.
\newblock


\bibitem[James(2017)]%
        {james2017impact}
\bibfield{author}{\bibinfo{person}{Lauren James}.}
  \bibinfo{year}{2017}\natexlab{}.
\newblock \showarticletitle{The Impact of Values as Heuristics on Social
  Cognition}.
\newblock  (\bibinfo{year}{2017}).
\newblock


\bibitem[Jones(2020)]%
        {jones2020cerebral}
\bibfield{author}{\bibinfo{person}{C Jones}.} \bibinfo{year}{2020}\natexlab{}.
\newblock \showarticletitle{The Cerebral Cortex Realizes a Universal
  Probabilistic Model of Computation in Complex Hilbert Spaces}.
\newblock  (\bibinfo{year}{2020}).
\newblock


\bibitem[Kahneman et~al\mbox{.}(1971)]%
        {kahneman1971belief}
\bibfield{author}{\bibinfo{person}{Daniel Kahneman} {et~al\mbox{.}}}
  \bibinfo{year}{1971}\natexlab{}.
\newblock \showarticletitle{Belief in the law of small numbers}.
\newblock \bibinfo{journal}{\emph{Psychological bulletin}}
  \bibinfo{volume}{76}, \bibinfo{number}{2} (\bibinfo{year}{1971}),
  \bibinfo{pages}{105--110}.
\newblock


\bibitem[Kaluarachchi et~al\mbox{.}(2021)]%
        {kaluarachchi2021review}
\bibfield{author}{\bibinfo{person}{Tharindu Kaluarachchi},
  \bibinfo{person}{Andrew Reis}, {and} \bibinfo{person}{Suranga Nanayakkara}.}
  \bibinfo{year}{2021}\natexlab{}.
\newblock \showarticletitle{A Review of Recent Deep Learning Approaches in
  Human-Centered Machine Learning}.
\newblock \bibinfo{journal}{\emph{Sensors}} \bibinfo{volume}{21},
  \bibinfo{number}{7} (\bibinfo{year}{2021}), \bibinfo{pages}{2514}.
\newblock


\bibitem[Kannan et~al\mbox{.}(2019)]%
        {kannan2019downstream}
\bibfield{author}{\bibinfo{person}{Sampath Kannan}, \bibinfo{person}{Aaron
  Roth}, {and} \bibinfo{person}{Juba Ziani}.} \bibinfo{year}{2019}\natexlab{}.
\newblock \showarticletitle{Downstream effects of affirmative action}. In
  \bibinfo{booktitle}{\emph{Proceedings of the Conference on Fairness,
  Accountability, and Transparency}}. \bibinfo{pages}{240--248}.
\newblock


\bibitem[Karlan(2021)]%
        {karlan2021reasoning}
\bibfield{author}{\bibinfo{person}{Brett Karlan}.}
  \bibinfo{year}{2021}\natexlab{}.
\newblock \showarticletitle{Reasoning with heuristics}.
\newblock \bibinfo{journal}{\emph{Ratio}} \bibinfo{volume}{34},
  \bibinfo{number}{2} (\bibinfo{year}{2021}), \bibinfo{pages}{100--108}.
\newblock


\bibitem[Ke et~al\mbox{.}(2018)]%
        {ke2018sparse}
\bibfield{author}{\bibinfo{person}{Nan~Rosemary Ke}, \bibinfo{person}{Anirudh
  Goyal}, \bibinfo{person}{Olexa Bilaniuk}, \bibinfo{person}{Jonathan Binas},
  \bibinfo{person}{Michael~C Mozer}, \bibinfo{person}{Chris Pal}, {and}
  \bibinfo{person}{Yoshua Bengio}.} \bibinfo{year}{2018}\natexlab{}.
\newblock \showarticletitle{Sparse attentive backtracking: Temporal
  creditassignment through reminding}.
\newblock \bibinfo{journal}{\emph{arXiv preprint arXiv:1809.03702}}
  (\bibinfo{year}{2018}).
\newblock


\bibitem[Kelly et~al\mbox{.}(2019)]%
        {kelly2019high}
\bibfield{author}{\bibinfo{person}{Matthew~A Kelly}, \bibinfo{person}{Nipun
  Arora}, \bibinfo{person}{Robert~L West}, {and} \bibinfo{person}{David
  Reitter}.} \bibinfo{year}{2019}\natexlab{}.
\newblock \showarticletitle{High-Dimensional Vector Spaces as the Architecture
  of Cognition.}. In \bibinfo{booktitle}{\emph{CogSci}}. \bibinfo{pages}{3491}.
\newblock


\bibitem[Kennedy(2012)]%
        {kennedy2012modelling}
\bibfield{author}{\bibinfo{person}{William~G Kennedy}.}
  \bibinfo{year}{2012}\natexlab{}.
\newblock \showarticletitle{Modelling human behaviour in agent-based models}.
\newblock In \bibinfo{booktitle}{\emph{Agent-based models of geographical
  systems}}. \bibinfo{publisher}{Springer}, \bibinfo{pages}{167--179}.
\newblock


\bibitem[Kerg et~al\mbox{.}(2020)]%
        {kerg2020untangling}
\bibfield{author}{\bibinfo{person}{Giancarlo Kerg}, \bibinfo{person}{Bhargav
  Kanuparthi}, \bibinfo{person}{Anirudh~Goyal ALIAS PARTH~GOYAL},
  \bibinfo{person}{Kyle Goyette}, \bibinfo{person}{Yoshua Bengio}, {and}
  \bibinfo{person}{Guillaume Lajoie}.} \bibinfo{year}{2020}\natexlab{}.
\newblock \showarticletitle{Untangling tradeoffs between recurrence and
  self-attention in artificial neural networks}.
\newblock \bibinfo{journal}{\emph{Advances in Neural Information Processing
  Systems}}  \bibinfo{volume}{33} (\bibinfo{year}{2020}).
\newblock


\bibitem[Khan et~al\mbox{.}(2020)]%
        {khan2020serious}
\bibfield{author}{\bibinfo{person}{Akif~Quddus Khan}, \bibinfo{person}{Salman
  Khan}, {and} \bibinfo{person}{Utkurbek Safaev}.}
  \bibinfo{year}{2020}\natexlab{}.
\newblock \showarticletitle{Serious Games and Gamification: A Systematic
  Literature Review}.
\newblock  (\bibinfo{year}{2020}).
\newblock


\bibitem[Kim(2021)]%
        {kim2021equilibrium}
\bibfield{author}{\bibinfo{person}{Jong~Gwang Kim}.}
  \bibinfo{year}{2021}\natexlab{}.
\newblock \showarticletitle{Equilibrium Computation of Generalized Nash Games:
  A New Lagrangian-Based Approach}.
\newblock \bibinfo{journal}{\emph{arXiv preprint arXiv:2106.00109}}
  (\bibinfo{year}{2021}).
\newblock


\bibitem[Klaproth et~al\mbox{.}(2019)]%
        {klaproth2019act}
\bibfield{author}{\bibinfo{person}{O Klaproth}, \bibinfo{person}{Marc
  Halbr{\"u}gge}, {and} \bibinfo{person}{Nele Russwinkel}.}
  \bibinfo{year}{2019}\natexlab{}.
\newblock \showarticletitle{ACT-R model for cognitive assistance in handling
  flight deck alerts}. In \bibinfo{booktitle}{\emph{Proceedings of the 17th
  International Conference on Cognitive Modeling, Montreal}}.
  \bibinfo{pages}{83--88}.
\newblock


\bibitem[Klaproth et~al\mbox{.}(2020)]%
        {klaproth2020neuroadaptive}
\bibfield{author}{\bibinfo{person}{Oliver~W Klaproth}, \bibinfo{person}{Marc
  Halbr{\"u}gge}, \bibinfo{person}{Laurens~R Krol}, \bibinfo{person}{Christoph
  Vernaleken}, \bibinfo{person}{Thorsten~O Zander}, {and} \bibinfo{person}{Nele
  Russwinkel}.} \bibinfo{year}{2020}\natexlab{}.
\newblock \showarticletitle{A Neuroadaptive Cognitive Model for Dealing With
  Uncertainty in Tracing Pilots' Cognitive State}.
\newblock \bibinfo{journal}{\emph{Topics in Cognitive Science}}
  \bibinfo{volume}{12}, \bibinfo{number}{3} (\bibinfo{year}{2020}),
  \bibinfo{pages}{1012--1029}.
\newblock


\bibitem[Kolekar et~al\mbox{.}(2021)]%
        {kolekar2021behavior}
\bibfield{author}{\bibinfo{person}{Suresh Kolekar}, \bibinfo{person}{Shilpa
  Gite}, \bibinfo{person}{Biswajeet Pradhan}, {and} \bibinfo{person}{Ketan
  Kotecha}.} \bibinfo{year}{2021}\natexlab{}.
\newblock \showarticletitle{Behavior Prediction of Traffic Actors for
  Intelligent Vehicle using Artificial Intelligence Techniques: A Review}.
\newblock \bibinfo{journal}{\emph{IEEE Access}} (\bibinfo{year}{2021}).
\newblock


\bibitem[Kotseruba and Tsotsos(2020)]%
        {kotseruba202040}
\bibfield{author}{\bibinfo{person}{Iuliia Kotseruba} {and}
  \bibinfo{person}{John~K Tsotsos}.} \bibinfo{year}{2020}\natexlab{}.
\newblock \showarticletitle{40 years of cognitive architectures: core cognitive
  abilities and practical applications}.
\newblock \bibinfo{journal}{\emph{Artificial Intelligence Review}}
  \bibinfo{volume}{53}, \bibinfo{number}{1} (\bibinfo{year}{2020}),
  \bibinfo{pages}{17--94}.
\newblock


\bibitem[Lakkaraju et~al\mbox{.}(2017)]%
        {lakkaraju2017selective}
\bibfield{author}{\bibinfo{person}{Himabindu Lakkaraju}, \bibinfo{person}{Jon
  Kleinberg}, \bibinfo{person}{Jure Leskovec}, \bibinfo{person}{Jens Ludwig},
  {and} \bibinfo{person}{Sendhil Mullainathan}.}
  \bibinfo{year}{2017}\natexlab{}.
\newblock \showarticletitle{The selective labels problem: Evaluating
  algorithmic predictions in the presence of unobservables}. In
  \bibinfo{booktitle}{\emph{Proceedings of the 23rd ACM SIGKDD International
  Conference on Knowledge Discovery and Data Mining}}.
  \bibinfo{pages}{275--284}.
\newblock


\bibitem[Lansdell et~al\mbox{.}(2019)]%
        {lansdell2019learning}
\bibfield{author}{\bibinfo{person}{Benjamin~James Lansdell},
  \bibinfo{person}{Prashanth~Ravi Prakash}, {and} \bibinfo{person}{Konrad~Paul
  Kording}.} \bibinfo{year}{2019}\natexlab{}.
\newblock \showarticletitle{Learning to solve the credit assignment problem}.
\newblock \bibinfo{journal}{\emph{arXiv preprint arXiv:1906.00889}}
  (\bibinfo{year}{2019}).
\newblock


\bibitem[Lau and Redlawsk(2001)]%
        {lau2001advantages}
\bibfield{author}{\bibinfo{person}{Richard~R Lau} {and}
  \bibinfo{person}{David~P Redlawsk}.} \bibinfo{year}{2001}\natexlab{}.
\newblock \showarticletitle{Advantages and disadvantages of cognitive
  heuristics in political decision making}.
\newblock \bibinfo{journal}{\emph{American Journal of Political Science}}
  (\bibinfo{year}{2001}), \bibinfo{pages}{951--971}.
\newblock


\bibitem[Lebiere(1999)]%
        {lebiere1999dynamics}
\bibfield{author}{\bibinfo{person}{Christian Lebiere}.}
  \bibinfo{year}{1999}\natexlab{}.
\newblock \showarticletitle{The dynamics of cognition: An ACT-R model of
  cognitive arithmetic}.
\newblock \bibinfo{journal}{\emph{Kognitionswissenschaft}} \bibinfo{volume}{8},
  \bibinfo{number}{1} (\bibinfo{year}{1999}), \bibinfo{pages}{5--19}.
\newblock


\bibitem[Lee(2007)]%
        {lee2007representation}
\bibfield{author}{\bibinfo{person}{Carole~J Lee}.}
  \bibinfo{year}{2007}\natexlab{}.
\newblock \showarticletitle{The representation of judgment heuristics and the
  generality problem}. In \bibinfo{booktitle}{\emph{Proceedings of the Annual
  Meeting of the Cognitive Science Society}}, Vol.~\bibinfo{volume}{29}.
\newblock


\bibitem[Lee(2018)]%
        {lee2018understanding}
\bibfield{author}{\bibinfo{person}{Min~Kyung Lee}.}
  \bibinfo{year}{2018}\natexlab{}.
\newblock \showarticletitle{Understanding perception of algorithmic decisions:
  Fairness, trust, and emotion in response to algorithmic management}.
\newblock \bibinfo{journal}{\emph{Big Data \& Society}} \bibinfo{volume}{5},
  \bibinfo{number}{1} (\bibinfo{year}{2018}),
  \bibinfo{pages}{2053951718756684}.
\newblock


\bibitem[Liao and He(2020)]%
        {liao2020racial}
\bibfield{author}{\bibinfo{person}{Yuting Liao} {and} \bibinfo{person}{Jiangen
  He}.} \bibinfo{year}{2020}\natexlab{}.
\newblock \showarticletitle{Racial mirroring effects on human-agent interaction
  in psychotherapeutic conversations}. In \bibinfo{booktitle}{\emph{Proceedings
  of the 25th International Conference on Intelligent User Interfaces}}.
  \bibinfo{pages}{430--442}.
\newblock


\bibitem[Lieder and Griffiths(2020)]%
        {lieder2020resource}
\bibfield{author}{\bibinfo{person}{Falk Lieder} {and} \bibinfo{person}{Thomas~L
  Griffiths}.} \bibinfo{year}{2020}\natexlab{}.
\newblock \showarticletitle{Resource-rational analysis: Understanding human
  cognition as the optimal use of limited computational resources}.
\newblock \bibinfo{journal}{\emph{Behavioral and Brain Sciences}}
  \bibinfo{volume}{43} (\bibinfo{year}{2020}).
\newblock


\bibitem[Lieder et~al\mbox{.}(2018)]%
        {lieder2018overrepresentation}
\bibfield{author}{\bibinfo{person}{Falk Lieder}, \bibinfo{person}{Thomas~L
  Griffiths}, {and} \bibinfo{person}{Ming Hsu}.}
  \bibinfo{year}{2018}\natexlab{}.
\newblock \showarticletitle{Overrepresentation of extreme events in decision
  making reflects rational use of cognitive resources.}
\newblock \bibinfo{journal}{\emph{Psychological review}} \bibinfo{volume}{125},
  \bibinfo{number}{1} (\bibinfo{year}{2018}), \bibinfo{pages}{1}.
\newblock


\bibitem[Lieder et~al\mbox{.}(2017)]%
        {lieder2017automatic}
\bibfield{author}{\bibinfo{person}{Falk Lieder}, \bibinfo{person}{Paul~M
  Krueger}, {and} \bibinfo{person}{Tom Griffiths}.}
  \bibinfo{year}{2017}\natexlab{}.
\newblock \showarticletitle{An automatic method for discovering rational
  heuristics for risky choice.}. In \bibinfo{booktitle}{\emph{CogSci}}.
\newblock


\bibitem[Lieto et~al\mbox{.}(2018)]%
        {lieto2018role}
\bibfield{author}{\bibinfo{person}{Antonio Lieto}, \bibinfo{person}{Mehul
  Bhatt}, \bibinfo{person}{Alessandro Oltramari}, {and} \bibinfo{person}{David
  Vernon}.} \bibinfo{year}{2018}\natexlab{}.
\newblock \bibinfo{title}{The role of cognitive architectures in general
  artificial intelligence}.
\newblock , \bibinfo{numpages}{3}~pages.
\newblock


\bibitem[Liu et~al\mbox{.}(2019)]%
        {liu2019self}
\bibfield{author}{\bibinfo{person}{Yang Liu}, \bibinfo{person}{Yifeng Zeng},
  \bibinfo{person}{Yingke Chen}, \bibinfo{person}{Jing Tang}, {and}
  \bibinfo{person}{Yinghui Pan}.} \bibinfo{year}{2019}\natexlab{}.
\newblock \showarticletitle{Self-improving generative adversarial reinforcement
  learning}. In \bibinfo{booktitle}{\emph{Proceedings of the 18th International
  Conference on Autonomous Agents and MultiAgent Systems}}.
  \bibinfo{pages}{52--60}.
\newblock


\bibitem[Lotz et~al\mbox{.}(2020)]%
        {lotz2020take}
\bibfield{author}{\bibinfo{person}{Alexander Lotz}, \bibinfo{person}{Nele
  Russwinkel}, {and} \bibinfo{person}{Enrico Wohlfarth}.}
  \bibinfo{year}{2020}\natexlab{}.
\newblock \showarticletitle{Take-over expectation and criticality in Level 3
  automated driving: a test track study on take-over behavior in semi-trucks}.
\newblock \bibinfo{journal}{\emph{Cognition, Technology \& Work}}
  \bibinfo{volume}{22}, \bibinfo{number}{4} (\bibinfo{year}{2020}),
  \bibinfo{pages}{733--744}.
\newblock


\bibitem[Mandelbaum et~al\mbox{.}(2020)]%
        {mandelbaum2020can}
\bibfield{author}{\bibinfo{person}{Eric Mandelbaum}, \bibinfo{person}{Isabel
  Won}, \bibinfo{person}{Steven Gross}, {and} \bibinfo{person}{Chaz
  Firestone}.} \bibinfo{year}{2020}\natexlab{}.
\newblock \showarticletitle{Can resources save rationality? ‘Anti-Bayesian’
  updating in cognition and perception}.
\newblock \bibinfo{journal}{\emph{Behavioral and Brain Sciences}}
  \bibinfo{volume}{143} (\bibinfo{year}{2020}).
\newblock


\bibitem[Marcot and Penman(2019)]%
        {marcot2019advances}
\bibfield{author}{\bibinfo{person}{Bruce~G Marcot} {and}
  \bibinfo{person}{Trent~D Penman}.} \bibinfo{year}{2019}\natexlab{}.
\newblock \showarticletitle{Advances in Bayesian network modelling: Integration
  of modelling technologies}.
\newblock \bibinfo{journal}{\emph{Environmental modelling \& software}}
  \bibinfo{volume}{111} (\bibinfo{year}{2019}), \bibinfo{pages}{386--393}.
\newblock


\bibitem[Milli et~al\mbox{.}(2021)]%
        {milli2021rational}
\bibfield{author}{\bibinfo{person}{Smitha Milli}, \bibinfo{person}{Falk
  Lieder}, {and} \bibinfo{person}{Thomas~L Griffiths}.}
  \bibinfo{year}{2021}\natexlab{}.
\newblock \showarticletitle{A rational reinterpretation of dual-process
  theories}.
\newblock \bibinfo{journal}{\emph{Cognition}}  \bibinfo{volume}{217}
  (\bibinfo{year}{2021}), \bibinfo{pages}{104881}.
\newblock


\bibitem[Mordacchini et~al\mbox{.}(2020)]%
        {mordacchini2020human}
\bibfield{author}{\bibinfo{person}{Matteo Mordacchini}, \bibinfo{person}{Marco
  Conti}, \bibinfo{person}{Andrea Passarella}, {and} \bibinfo{person}{Raffaele
  Bruno}.} \bibinfo{year}{2020}\natexlab{}.
\newblock \showarticletitle{Human-centric data dissemination in the IoP:
  Large-scale modeling and evaluation}.
\newblock \bibinfo{journal}{\emph{ACM Transactions on Autonomous and Adaptive
  Systems (TAAS)}} \bibinfo{volume}{14}, \bibinfo{number}{3}
  (\bibinfo{year}{2020}), \bibinfo{pages}{1--25}.
\newblock


\bibitem[Mordacchini et~al\mbox{.}(2017)]%
        {mordacchini2017social}
\bibfield{author}{\bibinfo{person}{Matteo Mordacchini}, \bibinfo{person}{Andrea
  Passarella}, {and} \bibinfo{person}{Marco Conti}.}
  \bibinfo{year}{2017}\natexlab{}.
\newblock \showarticletitle{A social cognitive heuristic for adaptive data
  dissemination in mobile Opportunistic Networks}.
\newblock \bibinfo{journal}{\emph{Pervasive and Mobile Computing}}
  \bibinfo{volume}{42} (\bibinfo{year}{2017}), \bibinfo{pages}{371--392}.
\newblock


\bibitem[Mordacchini et~al\mbox{.}(2016)]%
        {mordacchini2016design}
\bibfield{author}{\bibinfo{person}{Matteo Mordacchini},
  \bibinfo{person}{Lorenzo Valerio}, \bibinfo{person}{Marco Conti}, {and}
  \bibinfo{person}{Andrea Passarella}.} \bibinfo{year}{2016}\natexlab{}.
\newblock \showarticletitle{Design and evaluation of a cognitive approach for
  disseminating semantic knowledge and content in opportunistic networks}.
\newblock \bibinfo{journal}{\emph{Computer Communications}}
  \bibinfo{volume}{81} (\bibinfo{year}{2016}), \bibinfo{pages}{12--30}.
\newblock


\bibitem[Moreira et~al\mbox{.}(2019)]%
        {moreira2019towards}
\bibfield{author}{\bibinfo{person}{Catarina Moreira}, \bibinfo{person}{Lauren
  Fell}, \bibinfo{person}{Shahram Dehdashti}, \bibinfo{person}{Peter Bruza},
  {and} \bibinfo{person}{Andreas Wichert}.} \bibinfo{year}{2019}\natexlab{}.
\newblock \showarticletitle{Towards a quantum-like cognitive architecture for
  decision-making}.
\newblock \bibinfo{journal}{\emph{arXiv preprint arXiv:1905.05176}}
  (\bibinfo{year}{2019}).
\newblock


\bibitem[Moreira et~al\mbox{.}(2020)]%
        {moreira2020quantum}
\bibfield{author}{\bibinfo{person}{Catarina Moreira}, \bibinfo{person}{Prayag
  Tiwari}, \bibinfo{person}{Hari~Mohan Pandey}, \bibinfo{person}{Peter Bruza},
  {and} \bibinfo{person}{Andreas Wichert}.} \bibinfo{year}{2020}\natexlab{}.
\newblock \showarticletitle{Quantum-like influence diagrams for
  decision-making}.
\newblock \bibinfo{journal}{\emph{Neural Networks}}  \bibinfo{volume}{132}
  (\bibinfo{year}{2020}), \bibinfo{pages}{190--210}.
\newblock


\bibitem[Moreira and Wichert(2018)]%
        {moreira2018quantum}
\bibfield{author}{\bibinfo{person}{Catarina Moreira} {and}
  \bibinfo{person}{Andreas Wichert}.} \bibinfo{year}{2018}\natexlab{}.
\newblock \showarticletitle{Are quantum-like Bayesian networks more powerful
  than classical Bayesian networks?}
\newblock \bibinfo{journal}{\emph{Journal of Mathematical Psychology}}
  \bibinfo{volume}{82} (\bibinfo{year}{2018}), \bibinfo{pages}{73--83}.
\newblock


\bibitem[Morita et~al\mbox{.}(2020)]%
        {morita2020cognitive}
\bibfield{author}{\bibinfo{person}{Junya Morita}, \bibinfo{person}{Kazuhisa
  Miwa}, \bibinfo{person}{Akihiro Maehigashi}, \bibinfo{person}{Hitoshi Terai},
  \bibinfo{person}{Kazuaki Kojima}, {and} \bibinfo{person}{Frank~E Ritter}.}
  \bibinfo{year}{2020}\natexlab{}.
\newblock \showarticletitle{Cognitive Modeling of Automation Adaptation in a
  Time Critical Task}.
\newblock \bibinfo{journal}{\emph{Frontiers in Psychology}}
  \bibinfo{volume}{11} (\bibinfo{year}{2020}).
\newblock


\bibitem[Morrison et~al\mbox{.}(2021)]%
        {morrison2021social}
\bibfield{author}{\bibinfo{person}{Cecily Morrison}, \bibinfo{person}{Edward
  Cutrell}, \bibinfo{person}{Martin Grayson}, \bibinfo{person}{Anja Thieme},
  \bibinfo{person}{Alex Taylor}, \bibinfo{person}{Geert Roumen},
  \bibinfo{person}{Camilla Longden}, \bibinfo{person}{Sebastian Tschiatschek},
  \bibinfo{person}{Rita Faia~Marques}, {and} \bibinfo{person}{Abigail Sellen}.}
  \bibinfo{year}{2021}\natexlab{}.
\newblock \showarticletitle{Social Sensemaking with AI: Designing an Open-ended
  AI experience with a Blind Child}. In \bibinfo{booktitle}{\emph{Proceedings
  of the 2021 CHI Conference on Human Factors in Computing Systems}}.
  \bibinfo{pages}{1--14}.
\newblock


\bibitem[M{\"u}ller-Hansen et~al\mbox{.}(2017)]%
        {muller2017towards}
\bibfield{author}{\bibinfo{person}{Finn M{\"u}ller-Hansen},
  \bibinfo{person}{Maja Schl{\"u}ter}, \bibinfo{person}{Michael M{\"a}s},
  \bibinfo{person}{Jonathan~F Donges}, \bibinfo{person}{Jakob~J Kolb},
  \bibinfo{person}{Kirsten Thonicke}, {and} \bibinfo{person}{Jobst Heitzig}.}
  \bibinfo{year}{2017}\natexlab{}.
\newblock \showarticletitle{Towards representing human behavior and decision
  making in Earth system models--an overview of techniques and approaches}.
\newblock \bibinfo{journal}{\emph{Earth System Dynamics}} \bibinfo{volume}{8},
  \bibinfo{number}{4} (\bibinfo{year}{2017}), \bibinfo{pages}{977--1007}.
\newblock


\bibitem[Mutlu(2020)]%
        {mutlu2020quantum}
\bibfield{author}{\bibinfo{person}{Ece~C Mutlu}.}
  \bibinfo{year}{2020}\natexlab{}.
\newblock \showarticletitle{Quantum Probabilistic Models Using Feynman Diagram
  Rules for Better Understanding the Information Diffusion Dynamics in Online
  Social Networks}. In \bibinfo{booktitle}{\emph{Proceedings of the AAAI
  Conference on Artificial Intelligence}}, Vol.~\bibinfo{volume}{34}.
  \bibinfo{pages}{13730--13731}.
\newblock


\bibitem[Najar and Chetouani(2021)]%
        {najar2021reinforcement}
\bibfield{author}{\bibinfo{person}{Anis Najar} {and} \bibinfo{person}{Mohamed
  Chetouani}.} \bibinfo{year}{2021}\natexlab{}.
\newblock \showarticletitle{Reinforcement learning with human advice: a
  survey}.
\newblock \bibinfo{journal}{\emph{Frontiers in Robotics and AI}}
  \bibinfo{volume}{8} (\bibinfo{year}{2021}).
\newblock


\bibitem[Navidi and Landry(2021)]%
        {navidi2021new}
\bibfield{author}{\bibinfo{person}{Neda Navidi} {and} \bibinfo{person}{Rene
  Landry}.} \bibinfo{year}{2021}\natexlab{}.
\newblock \showarticletitle{New approach in human-AI interaction by
  reinforcement-imitation learning}.
\newblock \bibinfo{journal}{\emph{Applied Sciences}} \bibinfo{volume}{11},
  \bibinfo{number}{7} (\bibinfo{year}{2021}), \bibinfo{pages}{3068}.
\newblock


\bibitem[Nguyen and Gonzalez(2020a)]%
        {nguyen2020cognitive}
\bibfield{author}{\bibinfo{person}{Thuy~N Nguyen} {and}
  \bibinfo{person}{Cleotilde Gonzalez}.} \bibinfo{year}{2020}\natexlab{a}.
\newblock \bibinfo{booktitle}{\emph{Cognitive machine theory of mind}}.
\newblock \bibinfo{type}{{T}echnical {R}eport}. \bibinfo{institution}{Carnegie
  Mellon University}.
\newblock


\bibitem[Nguyen and Gonzalez(2020b)]%
        {nguyen2020effects}
\bibfield{author}{\bibinfo{person}{Thuy~N Nguyen} {and}
  \bibinfo{person}{Cleotilde Gonzalez}.} \bibinfo{year}{2020}\natexlab{b}.
\newblock \bibinfo{booktitle}{\emph{Effects of Decision Complexity in Goal
  seeking Gridworlds: A Comparison of Instance Based Learning and Reinforcement
  Learning Agents}}.
\newblock \bibinfo{type}{{T}echnical {R}eport}. \bibinfo{institution}{Carnegie
  Mellon University}.
\newblock


\bibitem[Ning et~al\mbox{.}(2021)]%
        {ning2021survey}
\bibfield{author}{\bibinfo{person}{Huansheng Ning}, \bibinfo{person}{Rui Yin},
  \bibinfo{person}{Ata Ullah}, {and} \bibinfo{person}{Feifei Shi}.}
  \bibinfo{year}{2021}\natexlab{}.
\newblock \showarticletitle{A Survey on Hybrid Human-Artificial Intelligence
  for Autonomous Driving}.
\newblock \bibinfo{journal}{\emph{IEEE Transactions on Intelligent
  Transportation Systems}} (\bibinfo{year}{2021}).
\newblock


\bibitem[Niv(2019)]%
        {niv2019learning}
\bibfield{author}{\bibinfo{person}{Yael Niv}.} \bibinfo{year}{2019}\natexlab{}.
\newblock \showarticletitle{Learning task-state representations}.
\newblock \bibinfo{journal}{\emph{Nature neuroscience}} \bibinfo{volume}{22},
  \bibinfo{number}{10} (\bibinfo{year}{2019}), \bibinfo{pages}{1544--1553}.
\newblock


\bibitem[Oaksford and Chater(2001)]%
        {oaksford2001probabilistic}
\bibfield{author}{\bibinfo{person}{Mike Oaksford} {and} \bibinfo{person}{Nick
  Chater}.} \bibinfo{year}{2001}\natexlab{}.
\newblock \showarticletitle{The probabilistic approach to human reasoning}.
\newblock \bibinfo{journal}{\emph{Trends in cognitive sciences}}
  \bibinfo{volume}{5}, \bibinfo{number}{8} (\bibinfo{year}{2001}),
  \bibinfo{pages}{349--357}.
\newblock


\bibitem[Oroojlooy et~al\mbox{.}(2020)]%
        {oroojlooy2020attendlight}
\bibfield{author}{\bibinfo{person}{Afshin Oroojlooy},
  \bibinfo{person}{Mohammadreza Nazari}, \bibinfo{person}{Davood Hajinezhad},
  {and} \bibinfo{person}{Jorge Silva}.} \bibinfo{year}{2020}\natexlab{}.
\newblock \showarticletitle{AttendLight: Universal Attention-Based
  Reinforcement Learning Model for Traffic Signal Control}.
\newblock \bibinfo{journal}{\emph{arXiv preprint arXiv:2010.05772}}
  (\bibinfo{year}{2020}).
\newblock


\bibitem[Pentecost et~al\mbox{.}(2016)]%
        {pentecost2016using}
\bibfield{author}{\bibinfo{person}{D Pentecost}, \bibinfo{person}{Charlotte
  Sennersten}, \bibinfo{person}{R Ollington}, \bibinfo{person}{C Lindley},
  {and} \bibinfo{person}{B Kang}.} \bibinfo{year}{2016}\natexlab{}.
\newblock \showarticletitle{Using a physics engine in ACT-R to aid decision
  making}.
\newblock \bibinfo{journal}{\emph{International Journal on Advances in
  Intelligent Systems}} \bibinfo{volume}{9}, \bibinfo{number}{3-4}
  (\bibinfo{year}{2016}), \bibinfo{pages}{298--309}.
\newblock


\bibitem[Peterson and Beach(1967)]%
        {peterson1967man}
\bibfield{author}{\bibinfo{person}{Cameron~R Peterson} {and}
  \bibinfo{person}{Lee~Roy Beach}.} \bibinfo{year}{1967}\natexlab{}.
\newblock \showarticletitle{Man as an intuitive statistician.}
\newblock \bibinfo{journal}{\emph{Psychological bulletin}}
  \bibinfo{volume}{68}, \bibinfo{number}{1} (\bibinfo{year}{1967}),
  \bibinfo{pages}{29}.
\newblock


\bibitem[Pothos and Busemeyer(2009)]%
        {pothos2009quantum}
\bibfield{author}{\bibinfo{person}{Emmanuel~M Pothos} {and}
  \bibinfo{person}{Jerome~R Busemeyer}.} \bibinfo{year}{2009}\natexlab{}.
\newblock \showarticletitle{A quantum probability explanation for violations of
  ‘rational’decision theory}.
\newblock \bibinfo{journal}{\emph{Proceedings of the Royal Society B:
  Biological Sciences}} \bibinfo{volume}{276}, \bibinfo{number}{1665}
  (\bibinfo{year}{2009}), \bibinfo{pages}{2171--2178}.
\newblock


\bibitem[Preuss et~al\mbox{.}(2019)]%
        {preuss2019implementation}
\bibfield{author}{\bibinfo{person}{Kai Preuss}, \bibinfo{person}{Leonie
  Raddatz}, {and} \bibinfo{person}{Nele Russwinkel}.}
  \bibinfo{year}{2019}\natexlab{}.
\newblock \showarticletitle{An implementation of Universal Spatial
  Transformative Cognition in ACT-R}. In \bibinfo{booktitle}{\emph{Proceedings
  of the 17th International Conference on Cognitive Modelling, TC Stewart
  (Ed.). University of Waterloo, Waterloo, Canada}}. \bibinfo{pages}{144--150}.
\newblock


\bibitem[Puig et~al\mbox{.}(2020)]%
        {puig2020watch}
\bibfield{author}{\bibinfo{person}{Xavier Puig}, \bibinfo{person}{Tianmin Shu},
  \bibinfo{person}{Shuang Li}, \bibinfo{person}{Zilin Wang},
  \bibinfo{person}{Yuan-Hong Liao}, \bibinfo{person}{Joshua~B Tenenbaum},
  \bibinfo{person}{Sanja Fidler}, {and} \bibinfo{person}{Antonio Torralba}.}
  \bibinfo{year}{2020}\natexlab{}.
\newblock \showarticletitle{Watch-and-help: A challenge for social perception
  and human-AI collaboration}.
\newblock \bibinfo{journal}{\emph{arXiv preprint arXiv:2010.09890}}
  (\bibinfo{year}{2020}).
\newblock


\bibitem[Qiu et~al\mbox{.}(2020)]%
        {qiu2020learning}
\bibfield{author}{\bibinfo{person}{Yiding Qiu}, \bibinfo{person}{Anwesan Pal},
  {and} \bibinfo{person}{Henrik~I Christensen}.}
  \bibinfo{year}{2020}\natexlab{}.
\newblock \showarticletitle{Learning hierarchical relationships for object-goal
  navigation}.
\newblock \bibinfo{journal}{\emph{arXiv preprint arXiv:2003.06749}}
  (\bibinfo{year}{2020}).
\newblock


\bibitem[Raghu et~al\mbox{.}(2019)]%
        {raghu2019algorithmic}
\bibfield{author}{\bibinfo{person}{Maithra Raghu}, \bibinfo{person}{Katy
  Blumer}, \bibinfo{person}{Greg Corrado}, \bibinfo{person}{Jon Kleinberg},
  \bibinfo{person}{Ziad Obermeyer}, {and} \bibinfo{person}{Sendhil
  Mullainathan}.} \bibinfo{year}{2019}\natexlab{}.
\newblock \showarticletitle{The algorithmic automation problem: Prediction,
  triage, and human effort}.
\newblock \bibinfo{journal}{\emph{arXiv preprint arXiv:1903.12220}}
  (\bibinfo{year}{2019}).
\newblock


\bibitem[Ramaraj et~al\mbox{.}(2021)]%
        {ramaraj2021unpacking}
\bibfield{author}{\bibinfo{person}{Preeti Ramaraj}, \bibinfo{person}{Charles~L
  Ortiz~Jr}, \bibinfo{person}{Matthew Klenk}, {and} \bibinfo{person}{Shiwali
  Mohan}.} \bibinfo{year}{2021}\natexlab{}.
\newblock \showarticletitle{Unpacking Human Teachers' Intentions For Natural
  Interactive Task Learning}.
\newblock \bibinfo{journal}{\emph{arXiv preprint arXiv:2102.06755}}
  (\bibinfo{year}{2021}).
\newblock


\bibitem[Ramaswamy et~al\mbox{.}(2018)]%
        {ramaswamy2018consistent}
\bibfield{author}{\bibinfo{person}{Harish~G Ramaswamy}, \bibinfo{person}{Ambuj
  Tewari}, {and} \bibinfo{person}{Shivani Agarwal}.}
  \bibinfo{year}{2018}\natexlab{}.
\newblock \showarticletitle{Consistent algorithms for multiclass classification
  with an abstain option}.
\newblock \bibinfo{journal}{\emph{Electronic Journal of Statistics}}
  \bibinfo{volume}{12}, \bibinfo{number}{1} (\bibinfo{year}{2018}),
  \bibinfo{pages}{530--554}.
\newblock


\bibitem[Rastogi et~al\mbox{.}(2020)]%
        {rastogi2020deciding}
\bibfield{author}{\bibinfo{person}{Charvi Rastogi}, \bibinfo{person}{Yunfeng
  Zhang}, \bibinfo{person}{Dennis Wei}, \bibinfo{person}{Kush~R Varshney},
  \bibinfo{person}{Amit Dhurandhar}, {and} \bibinfo{person}{Richard Tomsett}.}
  \bibinfo{year}{2020}\natexlab{}.
\newblock \showarticletitle{Deciding Fast and Slow: The Role of Cognitive
  Biases in AI-assisted Decision-making}.
\newblock \bibinfo{journal}{\emph{arXiv preprint arXiv:2010.07938}}
  (\bibinfo{year}{2020}).
\newblock


\bibitem[Reddy et~al\mbox{.}(2018)]%
        {reddy2018shared}
\bibfield{author}{\bibinfo{person}{Siddharth Reddy}, \bibinfo{person}{Anca~D
  Dragan}, {and} \bibinfo{person}{Sergey Levine}.}
  \bibinfo{year}{2018}\natexlab{}.
\newblock \showarticletitle{Shared autonomy via deep reinforcement learning}.
\newblock \bibinfo{journal}{\emph{arXiv preprint arXiv:1802.01744}}
  (\bibinfo{year}{2018}).
\newblock


\bibitem[Reverdy et~al\mbox{.}(2014)]%
        {reverdy2014modeling}
\bibfield{author}{\bibinfo{person}{Paul~B Reverdy}, \bibinfo{person}{Vaibhav
  Srivastava}, {and} \bibinfo{person}{Naomi~Ehrich Leonard}.}
  \bibinfo{year}{2014}\natexlab{}.
\newblock \showarticletitle{Modeling human decision making in generalized
  Gaussian multiarmed bandits}.
\newblock \bibinfo{journal}{\emph{Proc. IEEE}} \bibinfo{volume}{102},
  \bibinfo{number}{4} (\bibinfo{year}{2014}), \bibinfo{pages}{544--571}.
\newblock


\bibitem[Ritter et~al\mbox{.}(2019)]%
        {ritter2019act}
\bibfield{author}{\bibinfo{person}{Frank~E Ritter}, \bibinfo{person}{Farnaz
  Tehranchi}, {and} \bibinfo{person}{Jacob~D Oury}.}
  \bibinfo{year}{2019}\natexlab{}.
\newblock \showarticletitle{ACT-R: A cognitive architecture for modeling
  cognition}.
\newblock \bibinfo{journal}{\emph{Wiley Interdisciplinary Reviews: Cognitive
  Science}} \bibinfo{volume}{10}, \bibinfo{number}{3} (\bibinfo{year}{2019}),
  \bibinfo{pages}{e1488}.
\newblock


\bibitem[Rizun and Taranenko(2014)]%
        {rizun2014simulation}
\bibfield{author}{\bibinfo{person}{Nina Rizun} {and} \bibinfo{person}{Yurii
  Taranenko}.} \bibinfo{year}{2014}\natexlab{}.
\newblock \showarticletitle{Simulation models of human decision-making
  processes}.
\newblock \bibinfo{journal}{\emph{Management Dynamics in the Knowledge
  Economy}} \bibinfo{volume}{2}, \bibinfo{number}{2} (\bibinfo{year}{2014}),
  \bibinfo{pages}{241--264}.
\newblock


\bibitem[Russell and Norvig(2002)]%
        {russell2002artificial}
\bibfield{author}{\bibinfo{person}{Stuart Russell} {and} \bibinfo{person}{Peter
  Norvig}.} \bibinfo{year}{2002}\natexlab{}.
\newblock \showarticletitle{Artificial intelligence: a modern approach}.
\newblock  (\bibinfo{year}{2002}).
\newblock


\bibitem[Russwinkel et~al\mbox{.}(2018)]%
        {russwinkel2018act}
\bibfield{author}{\bibinfo{person}{Nele Russwinkel}, \bibinfo{person}{Sabine
  Prezenski}, \bibinfo{person}{Lisa D{\"o}rr}, {and} \bibinfo{person}{Frank
  Tamborello}.} \bibinfo{year}{2018}\natexlab{}.
\newblock \showarticletitle{ACT-Droid meets ACT-Touch: Modelling differences in
  swiping behavior with real Apps}. In \bibinfo{booktitle}{\emph{Proceedings of
  the 16th International Conference on Cognitive Modeling (ICCM 2018)}}.
  \bibinfo{pages}{21--24}.
\newblock


\bibitem[Samuelson(1995)]%
        {samuelson1995bounded}
\bibfield{author}{\bibinfo{person}{Larry Samuelson}.}
  \bibinfo{year}{1995}\natexlab{}.
\newblock \showarticletitle{Bounded rationality and game theory}.
\newblock \bibinfo{journal}{\emph{Quarterly Review of Economics and finance}}
  \bibinfo{volume}{36} (\bibinfo{year}{1995}), \bibinfo{pages}{17--36}.
\newblock


\bibitem[Scharfe and Russwinkel(2019a)]%
        {scharfe2019cognitive}
\bibfield{author}{\bibinfo{person}{Marlene Scharfe} {and} \bibinfo{person}{Nele
  Russwinkel}.} \bibinfo{year}{2019}\natexlab{a}.
\newblock \showarticletitle{A Cognitive Model for Understanding the Takeover in
  Highly Automated Driving Depending on the Objective Complexity of Non-Driving
  Related Tasks and the Traffic Environment.}. In
  \bibinfo{booktitle}{\emph{CogSci}}. \bibinfo{pages}{2734--2740}.
\newblock


\bibitem[Scharfe and Russwinkel(2019b)]%
        {scharfe2019towards}
\bibfield{author}{\bibinfo{person}{Marlene Scharfe} {and} \bibinfo{person}{Nele
  Russwinkel}.} \bibinfo{year}{2019}\natexlab{b}.
\newblock \showarticletitle{Towards a cognitive model of the takeover in highly
  automated driving for the improvement of human machine interaction}. In
  \bibinfo{booktitle}{\emph{TC Stewart (Chair), Proceedings of the 17th
  International Conference on cognitive modelling. Waterloo, Canada: University
  of Waterloo}}.
\newblock


\bibitem[Schneider(2020)]%
        {schneider2020humans}
\bibfield{author}{\bibinfo{person}{Johannes Schneider}.}
  \bibinfo{year}{2020}\natexlab{}.
\newblock \showarticletitle{Humans learn too: Better Human-AI Interaction using
  Optimized Human Inputs}.
\newblock \bibinfo{journal}{\emph{arXiv preprint arXiv:2009.09266}}
  (\bibinfo{year}{2020}).
\newblock


\bibitem[Schrimpf et~al\mbox{.}(2020)]%
        {schrimpf2020neural}
\bibfield{author}{\bibinfo{person}{Martin Schrimpf}, \bibinfo{person}{Idan
  Blank}, \bibinfo{person}{Greta Tuckute}, \bibinfo{person}{Carina Kauf},
  \bibinfo{person}{Eghbal~A Hosseini}, \bibinfo{person}{Nancy Kanwisher},
  \bibinfo{person}{Joshua Tenenbaum}, {and} \bibinfo{person}{Evelina
  Fedorenko}.} \bibinfo{year}{2020}\natexlab{}.
\newblock \showarticletitle{The neural architecture of language: Integrative
  reverse-engineering converges on a model for predictive processing}.
\newblock \bibinfo{journal}{\emph{BioRxiv}} (\bibinfo{year}{2020}).
\newblock


\bibitem[Simon(1990)]%
        {simon1990bounded}
\bibfield{author}{\bibinfo{person}{Herbert~A Simon}.}
  \bibinfo{year}{1990}\natexlab{}.
\newblock \showarticletitle{Bounded rationality}.
\newblock In \bibinfo{booktitle}{\emph{Utility and probability}}.
  \bibinfo{publisher}{Springer}, \bibinfo{pages}{15--18}.
\newblock


\bibitem[Singh et~al\mbox{.}(2018)]%
        {singh2018behavior}
\bibfield{author}{\bibinfo{person}{Meghendra Singh}, \bibinfo{person}{Achla
  Marathe}, \bibinfo{person}{Madhav~V Marathe}, {and} \bibinfo{person}{Samarth
  Swarup}.} \bibinfo{year}{2018}\natexlab{}.
\newblock \showarticletitle{Behavior model calibration for epidemic
  simulations}. In \bibinfo{booktitle}{\emph{Proceedings of the 17th
  International Conference on Autonomous Agents and MultiAgent Systems}}.
  \bibinfo{pages}{1640--1648}.
\newblock


\bibitem[Sinz et~al\mbox{.}(2019)]%
        {sinz2019engineering}
\bibfield{author}{\bibinfo{person}{Fabian~H Sinz}, \bibinfo{person}{Xaq
  Pitkow}, \bibinfo{person}{Jacob Reimer}, \bibinfo{person}{Matthias Bethge},
  {and} \bibinfo{person}{Andreas~S Tolias}.} \bibinfo{year}{2019}\natexlab{}.
\newblock \showarticletitle{Engineering a less artificial intelligence}.
\newblock \bibinfo{journal}{\emph{Neuron}} \bibinfo{volume}{103},
  \bibinfo{number}{6} (\bibinfo{year}{2019}), \bibinfo{pages}{967--979}.
\newblock


\bibitem[Smart et~al\mbox{.}(2016)]%
        {smart2016integrating}
\bibfield{author}{\bibinfo{person}{Paul~Richard Smart}, \bibinfo{person}{Tom
  Scutt}, \bibinfo{person}{Katia Sycara}, {and} \bibinfo{person}{Nigel~R
  Shadbolt}.} \bibinfo{year}{2016}\natexlab{}.
\newblock \showarticletitle{Integrating ACT-R cognitive models with the Unity
  game engine}.
\newblock In \bibinfo{booktitle}{\emph{Integrating cognitive architectures into
  virtual character design}}. \bibinfo{publisher}{IGI Global},
  \bibinfo{pages}{35--64}.
\newblock


\bibitem[Stocco et~al\mbox{.}(2021)]%
        {stocco2021analysis}
\bibfield{author}{\bibinfo{person}{Andrea Stocco}, \bibinfo{person}{Catherine
  Sibert}, \bibinfo{person}{Zoe Steine-Hanson}, \bibinfo{person}{Natalie Koh},
  \bibinfo{person}{John~E Laird}, \bibinfo{person}{Christian~J Lebiere}, {and}
  \bibinfo{person}{Paul Rosenbloom}.} \bibinfo{year}{2021}\natexlab{}.
\newblock \showarticletitle{Analysis of the human connectome data supports the
  notion of a “Common Model of Cognition” for human and human-like
  intelligence across domains}.
\newblock \bibinfo{journal}{\emph{NeuroImage}}  \bibinfo{volume}{235}
  (\bibinfo{year}{2021}), \bibinfo{pages}{118035}.
\newblock


\bibitem[Streicher et~al\mbox{.}(2021)]%
        {streicher2021dynamic}
\bibfield{author}{\bibinfo{person}{Alexander Streicher},
  \bibinfo{person}{Julius Busch}, {and} \bibinfo{person}{Wolfgang Roller}.}
  \bibinfo{year}{2021}\natexlab{}.
\newblock \showarticletitle{Dynamic Cognitive Modeling for Adaptive Serious
  Games}. In \bibinfo{booktitle}{\emph{International Conference on
  Human-Computer Interaction}}. Springer, \bibinfo{pages}{167--184}.
\newblock


\bibitem[Sun and Helie(2012)]%
        {sun2012reasoning}
\bibfield{author}{\bibinfo{person}{Ron Sun} {and} \bibinfo{person}{Sebastien
  Helie}.} \bibinfo{year}{2012}\natexlab{}.
\newblock \showarticletitle{Reasoning with heuristics and induction: An account
  based on the CLARION cognitive architecture}. In
  \bibinfo{booktitle}{\emph{The 2012 International Joint Conference on Neural
  Networks (IJCNN)}}. IEEE, \bibinfo{pages}{1--8}.
\newblock


\bibitem[Suomala(2020)]%
        {suomala2020consumer}
\bibfield{author}{\bibinfo{person}{Jyrki Suomala}.}
  \bibinfo{year}{2020}\natexlab{}.
\newblock \showarticletitle{The Consumer Contextual Decision-Making Model}.
\newblock \bibinfo{journal}{\emph{Frontiers in Psychology}}
  \bibinfo{volume}{11} (\bibinfo{year}{2020}), \bibinfo{pages}{2543}.
\newblock


\bibitem[Tenenbaum et~al\mbox{.}(2011)]%
        {tenenbaum2011grow}
\bibfield{author}{\bibinfo{person}{Joshua~B Tenenbaum},
  \bibinfo{person}{Charles Kemp}, \bibinfo{person}{Thomas~L Griffiths}, {and}
  \bibinfo{person}{Noah~D Goodman}.} \bibinfo{year}{2011}\natexlab{}.
\newblock \showarticletitle{How to grow a mind: Statistics, structure, and
  abstraction}.
\newblock \bibinfo{journal}{\emph{science}} \bibinfo{volume}{331},
  \bibinfo{number}{6022} (\bibinfo{year}{2011}), \bibinfo{pages}{1279--1285}.
\newblock


\bibitem[Thomson et~al\mbox{.}(2015)]%
        {thomson2015general}
\bibfield{author}{\bibinfo{person}{Robert Thomson}, \bibinfo{person}{Christian
  Lebiere}, \bibinfo{person}{John~R Anderson}, {and} \bibinfo{person}{James
  Staszewski}.} \bibinfo{year}{2015}\natexlab{}.
\newblock \showarticletitle{A general instance-based learning framework for
  studying intuitive decision-making in a cognitive architecture}.
\newblock \bibinfo{journal}{\emph{Journal of Applied Research in Memory and
  Cognition}} \bibinfo{volume}{4}, \bibinfo{number}{3} (\bibinfo{year}{2015}),
  \bibinfo{pages}{180--190}.
\newblock


\bibitem[Thulasidasan et~al\mbox{.}(2019)]%
        {thulasidasan2019combating}
\bibfield{author}{\bibinfo{person}{Sunil Thulasidasan}, \bibinfo{person}{Tanmoy
  Bhattacharya}, \bibinfo{person}{Jeff Bilmes}, \bibinfo{person}{Gopinath
  Chennupati}, {and} \bibinfo{person}{Jamal Mohd-Yusof}.}
  \bibinfo{year}{2019}\natexlab{}.
\newblock \showarticletitle{Combating label noise in deep learning using
  abstention}.
\newblock \bibinfo{journal}{\emph{arXiv preprint arXiv:1905.10964}}
  (\bibinfo{year}{2019}).
\newblock


\bibitem[Tversky and Kahneman(1974)]%
        {tversky1974judgment}
\bibfield{author}{\bibinfo{person}{Amos Tversky} {and} \bibinfo{person}{Daniel
  Kahneman}.} \bibinfo{year}{1974}\natexlab{}.
\newblock \showarticletitle{Judgment under uncertainty: Heuristics and biases}.
\newblock \bibinfo{journal}{\emph{science}} \bibinfo{volume}{185},
  \bibinfo{number}{4157} (\bibinfo{year}{1974}), \bibinfo{pages}{1124--1131}.
\newblock


\bibitem[Ullman and Tenenbaum(2020)]%
        {ullman2020bayesian}
\bibfield{author}{\bibinfo{person}{Tomer~D Ullman} {and}
  \bibinfo{person}{Joshua~B Tenenbaum}.} \bibinfo{year}{2020}\natexlab{}.
\newblock \showarticletitle{Bayesian models of conceptual development: Learning
  as building models of the world}.
\newblock \bibinfo{journal}{\emph{Annual Review of Developmental Psychology}}
  \bibinfo{volume}{2} (\bibinfo{year}{2020}), \bibinfo{pages}{533--558}.
\newblock


\bibitem[Urban and Schmidt(2001)]%
        {urban2001pecs}
\bibfield{author}{\bibinfo{person}{Christoph Urban} {and}
  \bibinfo{person}{Bernd Schmidt}.} \bibinfo{year}{2001}\natexlab{}.
\newblock \showarticletitle{PECS--agent-based modelling of human behaviour}. In
  \bibinfo{booktitle}{\emph{Emotional and Intelligent--The Tangled Knot of
  Social Cognition, AAAI Fall Symposium Series, North Falmouth, MA. www. or.
  unipassau. de/5/publik/urban/CUrban01. pdf}}.
\newblock


\bibitem[Vaswani et~al\mbox{.}(2017)]%
        {vaswani2017attention}
\bibfield{author}{\bibinfo{person}{Ashish Vaswani}, \bibinfo{person}{Noam
  Shazeer}, \bibinfo{person}{Niki Parmar}, \bibinfo{person}{Jakob Uszkoreit},
  \bibinfo{person}{Llion Jones}, \bibinfo{person}{Aidan~N Gomez},
  \bibinfo{person}{{\L}ukasz Kaiser}, {and} \bibinfo{person}{Illia
  Polosukhin}.} \bibinfo{year}{2017}\natexlab{}.
\newblock \showarticletitle{Attention is all you need}.
\newblock \bibinfo{journal}{\emph{Advances in neural information processing
  systems}}  \bibinfo{volume}{30} (\bibinfo{year}{2017}).
\newblock


\bibitem[Wang et~al\mbox{.}(2013)]%
        {wang2013invisible}
\bibfield{author}{\bibinfo{person}{Chen Wang}, \bibinfo{person}{Pablo Cesar},
  {and} \bibinfo{person}{Erik Geelhoed}.} \bibinfo{year}{2013}\natexlab{}.
\newblock \showarticletitle{An Invisible Gorilla: Is It a Matter of Focus of
  Attention?}. In \bibinfo{booktitle}{\emph{Pacific-Rim Conference on
  Multimedia}}. Springer, \bibinfo{pages}{318--326}.
\newblock


\bibitem[Wei et~al\mbox{.}(2019)]%
        {wei2019presslight}
\bibfield{author}{\bibinfo{person}{Hua Wei}, \bibinfo{person}{Chacha Chen},
  \bibinfo{person}{Guanjie Zheng}, \bibinfo{person}{Kan Wu},
  \bibinfo{person}{Vikash Gayah}, \bibinfo{person}{Kai Xu}, {and}
  \bibinfo{person}{Zhenhui Li}.} \bibinfo{year}{2019}\natexlab{}.
\newblock \showarticletitle{Presslight: Learning max pressure control to
  coordinate traffic signals in arterial network}. In
  \bibinfo{booktitle}{\emph{Proceedings of the 25th ACM SIGKDD International
  Conference on Knowledge Discovery \& Data Mining}}.
  \bibinfo{pages}{1290--1298}.
\newblock


\bibitem[Whitehill(2013)]%
        {whitehill2013understanding}
\bibfield{author}{\bibinfo{person}{Jacob Whitehill}.}
  \bibinfo{year}{2013}\natexlab{}.
\newblock \showarticletitle{Understanding ACT-R-an outsider's perspective}.
\newblock \bibinfo{journal}{\emph{arXiv preprint arXiv:1306.0125}}
  (\bibinfo{year}{2013}).
\newblock


\bibitem[Wu et~al\mbox{.}(2020)]%
        {wu2020joint}
\bibfield{author}{\bibinfo{person}{Guojun Wu}, \bibinfo{person}{Yanhua Li},
  \bibinfo{person}{Shikai Luo}, \bibinfo{person}{Ge Song},
  \bibinfo{person}{Qichao Wang}, \bibinfo{person}{Jing He},
  \bibinfo{person}{Jieping Ye}, \bibinfo{person}{Xiaohu Qie}, {and}
  \bibinfo{person}{Hongtu Zhu}.} \bibinfo{year}{2020}\natexlab{}.
\newblock \showarticletitle{A Joint Inverse Reinforcement Learning and Deep
  Learning Model for Drivers' Behavioral Prediction}. In
  \bibinfo{booktitle}{\emph{Proceedings of the 29th ACM International
  Conference on Information \& Knowledge Management}}.
  \bibinfo{pages}{2805--2812}.
\newblock


\bibitem[Yang et~al\mbox{.}(2020)]%
        {yang2020re}
\bibfield{author}{\bibinfo{person}{Qian Yang}, \bibinfo{person}{Aaron
  Steinfeld}, \bibinfo{person}{Carolyn Ros{\'e}}, {and} \bibinfo{person}{John
  Zimmerman}.} \bibinfo{year}{2020}\natexlab{}.
\newblock \showarticletitle{Re-examining whether, why, and how human-AI
  interaction is uniquely difficult to design}. In
  \bibinfo{booktitle}{\emph{Proceedings of the 2020 chi conference on human
  factors in computing systems}}. \bibinfo{pages}{1--13}.
\newblock


\bibitem[Zhao et~al\mbox{.}(2019)]%
        {zhao2019explicit}
\bibfield{author}{\bibinfo{person}{Guangxiang Zhao}, \bibinfo{person}{Junyang
  Lin}, \bibinfo{person}{Zhiyuan Zhang}, \bibinfo{person}{Xuancheng Ren},
  \bibinfo{person}{Qi Su}, {and} \bibinfo{person}{Xu Sun}.}
  \bibinfo{year}{2019}\natexlab{}.
\newblock \showarticletitle{Explicit sparse transformer: Concentrated attention
  through explicit selection}.
\newblock \bibinfo{journal}{\emph{arXiv preprint arXiv:1912.11637}}
  (\bibinfo{year}{2019}).
\newblock


\bibitem[Zhu and Harrell(2008)]%
        {zhu2008daydreaming}
\bibfield{author}{\bibinfo{person}{Jichen Zhu} {and} \bibinfo{person}{D~Fox
  Harrell}.} \bibinfo{year}{2008}\natexlab{}.
\newblock \showarticletitle{Daydreaming with Intention: Scalable Blending-Based
  Imagining and Agency in Generative Interactive Narrative.}. In
  \bibinfo{booktitle}{\emph{AAAI Spring Symposium: Creative Intelligent
  Systems}}, Vol.~\bibinfo{volume}{156}.
\newblock


\bibitem[Zhu et~al\mbox{.}(2020)]%
        {zhu2020bayesian}
\bibfield{author}{\bibinfo{person}{Jian-Qiao Zhu}, \bibinfo{person}{Adam~N
  Sanborn}, {and} \bibinfo{person}{Nick Chater}.}
  \bibinfo{year}{2020}\natexlab{}.
\newblock \showarticletitle{The Bayesian sampler: Generic Bayesian inference
  causes incoherence in human probability judgments.}
\newblock \bibinfo{journal}{\emph{Psychological review}} \bibinfo{volume}{127},
  \bibinfo{number}{5} (\bibinfo{year}{2020}), \bibinfo{pages}{719}.
\newblock


\end{thebibliography}

\end{document}